\documentclass[preprint,3p]{elsarticle}  % add "twocolumn" here
\usepackage[table,xcdraw]{xcolor}
\usepackage[ruled,linesnumbered]{algorithm2e} 
\usepackage{booktabs}
\makeatletter
\renewcommand*{\@algocf@pre@ruled}{\hrule height\heavyrulewidth depth0pt \kern\belowrulesep}
\renewcommand*{\algocf@caption@ruled}{\box\algocf@capbox\kern\aboverulesep\hrule height\lightrulewidth\kern\belowrulesep}
\renewcommand*{\@algocf@post@ruled}{\kern\aboverulesep\hrule height\heavyrulewidth\relax}
\makeatother
\newcommand\labelAndRemember[2]
  {\expandafter\gdef\csname labeled:#1\endcsname{#2}%
   \label{#1}#2}
\newcommand\recallLabel[1]
   {\csname labeled:#1\endcsname\tag{\ref{#1}}}

\usepackage[skins]{tcolorbox}
\usepackage{makecell}
\usepackage{times}
\usepackage[hyphens]{url}  % use urls, break them on hyphens (and slashes ofc)
\usepackage{latexsym}
\usepackage{hyperref}
\usepackage{amssymb}
\usepackage{todonotes}
\usepackage{amsfonts}
\usepackage{amsmath}
\usepackage{xcolor}
\usepackage{soul}
\usepackage[toc,page]{appendix}
\usepackage{pythonhighlight}
\usepackage[symbol]{footmisc}
\lstnewenvironment{python_nobox}[1][]{\lstset{style=mypython, frame=none, #1}}{}

\begin{document}
\begin{frontmatter}

\title{Calculating and Visualizing Counterfactual Feature Importance Values}

\author[1,2]{Bjorge Meulemeester\footnote[1]{Authors contributed equally.}}
\ead{bjorgemeulemeester@mpinb.mpg.de}
\author[1]{Raphael Mazzine Barbosa De Oliveira$^*$}
\ead{Raphael.MazzineBarbosaDeOliveira@uantwerpen.be}
\author[1]{David Martens}
\ead{david.martens@uantwerpen.be}
\address[1]{Department of Engineering Management, University of Antwerp, Prinsstraat 13, 2000 Antwerpen, BEL}
\address[2]{Max-Planck-Institut für Neurobiologie des Verhaltens – Caesar, Ludwig-Erhard-Allee 2, 53175 Bonn, DE}

\begin{abstract}
Despite the success of complex machine learning algorithms, mostly justified by an outstanding performance in prediction tasks, their inherent opaque nature still represents a challenge to their responsible application. Counterfactual explanations surged as one potential solution to explain individual decision results. However, two major drawbacks directly impact their usability: (1) the isonomic view of feature changes, in which it is not possible to observe \textit{how much} each modified feature influences the prediction, and (2) the lack of graphical resources to visualize the counterfactual explanation. We introduce Counterfactual Feature (change) Importance (CFI) values as a solution: a way of assigning an importance value to each feature change in a given counterfactual explanation. To calculate these values, we propose two potential CFI methods. One is simple, fast, and has a greedy nature. The other, coined CounterShapley, provides a way to calculate Shapley values between the factual-counterfactual pair. Using these importance values, we additionally introduce three chart types to visualize the counterfactual explanations: (a) the Greedy chart, which shows a greedy sequential path for prediction score increase up to predicted class change, (b) the CounterShapley chart, depicting its respective score in a simple and one-dimensional chart, and finally (c) the Constellation chart, which shows all possible combinations of feature changes, and their impact on the model's prediction score. For each of our proposed CFI methods and visualization schemes, we show how they can provide more information on counterfactual explanations. Finally, an open-source implementation is offered, compatible with any counterfactual explanation generator algorithm.

\end{abstract}


\end{frontmatter}

\section{Introduction}
Artificial intelligence (AI) has become capable of capturing, learning, and predicting complex data structures. The performance of GPT-3~\cite{brown2020language} models, for example, are so performant that a piece of software can mimic language to a level that is often indistinguishable from the human-written text. Deep learning models can learn a voice's intonation, tone, and pronunciation habits using just a couple of seconds of audio as training data~\cite{ronanki2016template,neekhara2021expressive,yang2019self}.
Learning highly complex data structures, such as speech, generally requires an equally complex model to capture the nuances of the data accurately. While this increase in complexity opens up the possibility of improved performance, it comes with a price: its explainability is greatly reduced~\cite{arrieta2020explainable}. This lack of justification for decisions can decisively impact their application~\cite{meske2022explainable}. Current legislation, for example, requires explainability of models that impact high-stake decisions for people's lives~\cite{wachter2017counterfactual}. Moreover, understanding why a model works can provide insights to guide improvements in itself~\cite{arrieta2020explainable} and possibly detect erratic patterns learned from data that can cause discrimination or unfair decisions~\cite{goethals2022precof,martensdsethics2022,arrieta2020explainable}. Models that become inexplicable due to their inherent complex characteristics, such as a high number of parameters or complex feature associations, are referred to as ``black box'' models, since their inner workings are not clear enough for humans to comprehend.

The field of eXplainable Artificial Intelligence (XAI) dedicates great efforts to shed light on how complex models reach decisions by adopting several methods~\cite{speith2022review,martensdsethics2022}. 
The wide variety of XAI methods can generally be categorized by the scope of their explanation, model compatibility, and output format~\cite{arrieta2020explainable,speith2022review}. In terms of scope, local explanation methods aim at uncovering why a model predicts some outcome for a single data instance, while global explanations seek to provide an overall explanation of how the model works on the entire dataset. For model compatibility, we have methods that only work for a specific kind of model - since they often use particular mechanisms to generate explanations - and approaches that are independent of model type i.e. ``model-agnostic". Finally, the output format of the explanations can be simple feature importance values or more sophisticated representations such as heatmaps for image explanations.

Popular global explanation approaches include, but are not limited to: using intrinsic model characteristics, such as impurity decrease analysis for tree-based models~\cite{breiman2001random}; rule extraction strategies, such as Trepan~\cite{craven1996extracting} and ALBA~\cite{martensalba2009}, where a less complex but comprehensible surrogate model is built to mimic the black-box model, and its intrinsic explainable parts are used as a proxy to explain the black-box model; and depicting the marginal effect of features on the prediction score (partial dependence plots, PDP)~\cite{hastie01statisticallearning}. As for local explanations, LIME~\cite{lime} and SHAP~\cite{lundberg2017unified} are popular feature importance (FI) methods, indicating the most important (subset of) features for a given instance's prediction score. LIME considers data in the vicinity of the instance to be explained and fits a locally linear relationship. The coefficients of this linear fit can be interpreted as the FI values. SHAP calculates approximate or exact Shapley values, compared to the average prediction and average feature value of the entire dataset. Finally, counterfactual explanations provides a set of changes in feature values that lead to a different model decision~\cite{verma2020counterfactual}. %LIME and 

In terms of explanation output format, all listed explanation methods (except for counterfactual explanations) include ways to visualize the feature's importance graphically. LIME allows bar charts to visualize the weights of the locally linear decision boundary\footnote{The visualization resources for LIME can be seen at \href{https://github.com/marcotcr/lime}{https://github.com/marcotcr/lime}}. SHAP even allows various visual methods to compare the feature Shapley values, such as waterfall plots, force plots, the popular beeswarm and violin plots, and even image heatmaps for image recognition models\footnote{SHAP's charts can be found in their original repository: 
\href{https://github.com/slundberg/shap}{https://github.com/slundberg/shap}}. And the random forest's mean decrease in impurity (MDI) allows a simple representation of the feature's importance by bar charts~\cite{sklearn_api}.

One naive visualization approach that is commonly used to introduce counterfactual explanations is to show the factual and counterfactual points graphically in their input space. However, this is only possible for 2 to 3 dimensions, a rare occurrence for real-life datasets. Moreover, just showing the value changes in Cartesian space does not add considerable value to the explanation, since features usually have vastly different and interdependent distributions. They cannot be directly compared. Therefore, the most practical and adopted solution to represent the output of a counterfactual explanation method is to provide a set of feature names and associated values to be changed. We can illustrate such an explanation with the example below, which considers a machine learning model that approves or rejects credit card applications: 

\begin{center}
\begin{tcolorbox}[enhanced,width=10cm,center upper,drop fuzzy shadow southwest,
    boxrule=0.4pt,sharp corners,colframe=yellow!80!black,colback=yellow!10]
\textbf{Explanation 1: Counterfactual explanation why a credit card application was rejected. }
\noindent\rule{\textwidth}{0.4pt}
If your age were 28 years old instead of 20, and \\ your salary were \$4,000 instead of \$1,200\\ 
then you would be approved instead of
rejected.
\end{tcolorbox}
\end{center}

The representation does not include any visualization, nor any way to assess which of the feature changes was more important than the other. This is in stark contrast to LIME or SHAP, who have easily interpretable charts that show the importance of each feature.
Additionally, the unique objective of counterfactuals explanations is to look at feature \textit{changes}, rather than just features, that lead to a \textit{class change}. This does not allow for a simple translation of LIME or SHAP visualization.
Hence, this leads to the following two questions: can we obtain importance values for each feature change in a counterfactual explanation, and can these be used to visualize the explanation?

\begin{figure}[h]
\centering
\includegraphics[width=14cm]{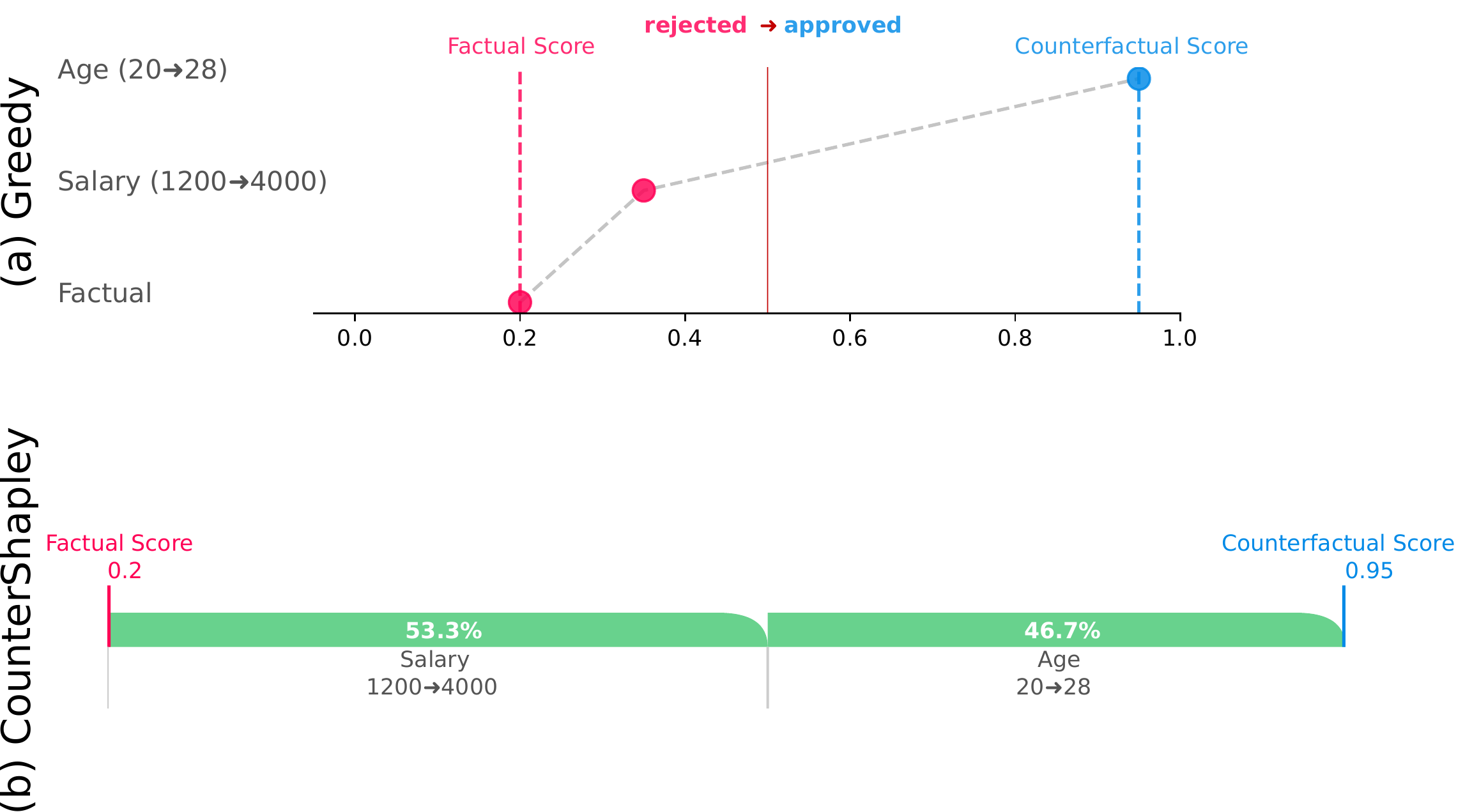}
\caption{Counterfactual explanation visualizations for \textit{explanation 1} using the (a) Greedy and (b) CounterShapley charts.
}
\label{fig:examplexai}
\end{figure}

In this paper, we solve this problem as follows: we generate Counterfactual Feature change Importance (CFI) values, and propose two methods to do so\footnote{Note that we will interchangeably talk about feature changes or feature value changes, as well as their associated CFI values and importance scores. Each feature change has a value in feature space, and an associated CFI value as well.}. 
Figure~\ref{fig:examplexai} already illustrates the goal and result of such CFI methods for the same counterfactual explanation we previously mentioned.
The \textit{Greedy CFI method and chart (a)}  
visualize the modification sequence that occurs when changing the feature with the greatest effect on the model prediction score, one at a time.
Since each change adapts the feature, each feature change will have a different effect after each modification, thus making these values dependent on the order of the feature changes.
The benefit of this approach is its efficiency and simplicity. In this example, we see that the prediction score for the data instance (the factual) is 0.2, which is below the 0.5 threshold, leading to a rejected decision from the model. Changing salary from 1,200 to 1,400 apparently leads to the highest increase in the predicted score, yet still leading to a rejection score. Additionally changing the age from 20 to 28 then leads to the counterfactual instance with a prediction score of 0.95 and approval decision. Note that changing the age only has an impact this high \textit{after} the salary has been changed as well.

The \textit{CounterShapley CFI method and chart} generalizes this concept, by looking at all possible subsets of feature changes, and assessing the average marginal value of each feature change. This naturally leads to Shapley values~\cite{shapley1953quota}. 
Section~\ref{sec:visualizingCFE} will describe this in more detail, but the example already indicates that salary change has an importance value of 53.3\%, while age is slightly less important, having an importance value of 46.7\%. Finally, we will also introduce \textit{Constellation charts}, that aim to visualize all these subsets of feature-value changes. Further demonstration of this concept will be provided in our empirical section. 

\section{Motivation}
\label{sec:motivation}

The research interest in counterfactual explanations is evident when we look at the increasing number of publications in this field~\cite{stepin2021survey}: ever since it has been introduced in 2011~\cite{martens2011explaining}, over 300 counterfactual generation methods have been proposed~\cite{verma2020counterfactual}. This interest is justified by several desirable properties that counterfactual explanations have: they are simple, model-agnostic, sparse, and decision-oriented~\cite{verma2020counterfactual,de2021framework,fernandez2020explaining}. The simplicity is not only related to the output explanation itself, but also refers to the familiarity that humans have to use counterfactual arguments to provide explanations~\cite{byrne2007precis}. This process is well-established both in philosophy and psychology~\cite{lipton1990contrastive,byrne2007precis}, showing that such explanations come natural to end users.
In theory, counterfactual explanations do not rely on any specific model since they consider the prediction model as a black box~\cite{stepin2021survey}, making them model-agnostic.
The explanations are typically also sparse, in the sense that only a small subset of all features occurs in the explanation. Not only does this make the explanation smaller and hence more comprehensible~\cite{martensdsethics2022} (compared to an explanation using all features), but it also considerably reduces the complexity of the problem of assigning importance values to each feature change in the explanation. This sparsity property thereby removes the need to limit the search or visualization to some smaller subset of features, as is done in LIME and SHAP where the number of features is user-defined.

Finally, the decision-oriented nature is a key point of counterfactuals; they focus on modifications that lead to a prediction change~\cite{fernandez2020explaining}. This characteristic is fundamental for highly complex machine learning methods, since their nonlinearity can lead to nontrivial feature-value relationships that can potentially bring misleading assumptions. We can verify this last point in literature works~\cite{vermeire2022explainable,fernandez2020explaining} that compare counterfactual explanations with feature importance-based methods (LIME and SHAP) and show that the most important features do not always lead to a change in classification decision. Hence, counterfactual explanations can be advantageous whenever one is more interested in explaining a predicted decision, rather than a prediction. This decision-oriented nature of counterfactuals aligns well with the transparency requirements in legislations, such as the European GDPR~\cite{wachter2017counterfactual}.

Despite these numerous benefits of counterfactual explanations, one major drawback, as introduced in the previous section, is the way these explanation results are presented. This issue is not only limited to the lack of visual approaches to plot charts, but also to the inability to assign meaningful values to feature changes. One might wonder: why not simply apply an existing feature importance method, such as SHAP, to the counterfactual explanation? The reason is multifaceted, as we will describe in detail in the next sections: (1) we focus on feature-value \textit{changes} (e.g. age changing from 20 to 28), not just on feature values (e.g. age equals 20); (2) the dimensionality that we have to deal with is generally much smaller, yielding unique visualisation opportunities; (3) the sum of an instance's SHAP values sum up to the difference between the average prediction score and the instance prediction, not to the difference between counterfactual and factual scores; and (4) the reference point and the features considered in SHAP are different from those considered in counterfactual explanations. So, we aim to define Counterfactual Feature change Importance (CFI) methods that enhance the informative value of counterfactual explanations by creating feature change importance values and correspondent visualization resources that are tailored to the cause.

The significance of visualization is well known in the field~\cite{verma2020counterfactual}, and it is further reinforced in a recent study in which consumers preferred feature importance methods over counterfactual explanations~\cite{ramon2021understanding}. The authors of the study conjecture that this preference is likely due to the lack of visualization for counterfactual explanations. Likewise, merely providing what feature values should be changed in order to generate a counterfactual explanation does not show any distinction in their relative importance. This additional information can provide useful information on how the model prediction score works, ultimately leading to a better comprehension of the model's decision. This would allow users and practitioners to critically evaluate whether the model behaves as expected.

The graphical representation of counterfactuals traces back to its origins~\cite{martens2014explaining}, where the greedy nature of the proposed counterfactual generation algorithm (SEDC) allowed the plotting of sequential changes until the counterfactual class is achieved. In more recent studies, the visualization problem was tackled by multiple studies. For example, GAMUT~\cite{hohman2019gamut} implements a user interface that shows multiple statistics and methods to assess the relationship between the instance's features and the model's prediction. Here, counterfactual visualisation is limited to a simple model output change given certain modifications, without any assignment of how important each of those changes is.
Google's What-If Tool~\cite{wexler2019if} focuses on allowing the user to explore the model's behavior by probing feature changes that are dynamically presented in tables and charts. But in terms of reporting, the counterfactual feature changes do not offer much information about their importance: they are depicted by simple tables that highlight the value changes, and two-dimensional charts with features values or different model's prediction scores.
DECE~\cite{cheng2020dece} also presents a user interface that includes various statistical analyses of features, and shows multiple counterfactual explanations according to sequential modification steps. But it still does not show the importance of each feature change to prediction scoring.
Lastly, VICE~\cite{gomez2020vice} represents counterfactual changes in charts together with the data distribution, and it allows customization of which features to include in the explanation. Yet it again does not disclose the influence of each feature change.
Although the previous approaches have advantages by allowing users to investigate interactively how counterfactuals work, and they include multiple statistical analyses related to the target explanation and its features, they introduce an extra layer of complexity that users must learn to operate. Furthermore, they all represent counterfactual changes by only showing what features were modified in tables or spatial shifts with two-dimensional charts. This all emphasizes the current lack of a method to represent how each counterfactual explanation feature change impacts the model's prediction scoring.

\section{Counterfactual Explanations}
\label{sec:counterfactual_explanations}
The methods presented here do not rely on any specific type of counterfactual generator. A CFI method simply assumes some counterfactual explanation has been provided with corresponding feature changes, for which importance values need to be assigned, irrespective of the algorithm that was used to generate the explanation. Therefore, this section introduces a general concept of counterfactual explanation, which covers any generation algorithm. The formal definition is then used to introduce the CFI methods' theoretical reasoning.

Counterfactual explanations~\cite{martens2014explaining,angelov2021explainable} assume four basic components: a dataset ($\textbf{D}$); an instance to explain ($\textbf{x}$); a prediction model ($\mathcal{M}$); and, in the case of a classification task, a threshold ($t$).

Consider an $N \times M$ dimension dataset, with a set of $N$ instances

\begin{equation}
    \textbf{D} = [\textbf{d}^1, \textbf{d}^2, ..., \textbf{d}^N] 
\end{equation}

and $M$ features
\begin{equation}
    \textbf{d}^n = [d^n_1, d^n_2, ..., d^n_M]\ ,\  n \in [1...N]
\end{equation}
where $\textbf{d}^n$ is called a datapoint or an instance: a vector of size $M$.

These instances are used by a prediction model $\mathcal{M}$ to assign a score $s$. Scoring an instance with a model will be denoted by
\begin{equation}
    \mathcal{M}(\textbf{d}^n) = s
\end{equation}

If we consider a binary classification task where we set a decision threshold $t$, assigning instances to \textit{class 0} if $s < t$ and to \textit{class 1} if $s \geq t$, we can effectively categorize all instances into either class. We will only assume binary classification in this work for simplicity, but this can easily be generalized to multiclass classification with, e.g., a one-vs-one or one-vs-rest paradigm. Since CFI only assumes the existence of some counterfactual and its associated feature values, no matter the prediction task, this can also be generalized to regression tasks.

Given two different datapoints $\boldsymbol{a}$ and $\boldsymbol{b}$, we can define the difference between $\boldsymbol{b}$ and $\boldsymbol{a}$ as the set of indices where their feature values differ:
\begin{equation}
\label{eq:set-indices}
    \delta_{a,b} = \{ i\ |\ a_i \neq b_i \}
\end{equation}

For notational ease, we define the scoring of the instance $\boldsymbol{a}$ with replaced feature values $\delta_{a,b}$ as being: 

\begin{equation}
    \mathcal{M}(\delta_{a,b}) = s
\end{equation}

Note that the definition of $\delta$ is symmetric with respect to the two associated instances. By convention, this notation will denote the scoring of $\textbf{a}$, adapted to have the values of $\textbf{b}$ on the indices in $\delta$, where $\textbf{a}$ and $\textbf{b}$ are the left and right instance of the $\neq$ sign in Equation~\ref{eq:set-indices}. This becomes important as soon as we consider subsets of $\delta_{a,b}$. In the case of factual and counterfactual instances (see below), this notation will always denote the factual instance, adapted to have feature values of the counterfactual on some indices. Not the other way around.

Using this notation, we can formalize a counterfactual instance $\textbf{c}$, derived from a factual instance $\textbf{x}\in \textbf{D}$, having the following properties:
\begin{enumerate}
    \item The factual and counterfactual's predicted classes are not the same:
    \begin{equation}
        \left(\mathcal{M}(\textbf{x}) > t\right) \neq \left(\mathcal{M}(\textbf{c}) > t\right)
    \end{equation}

    \item The difference between the factual and counterfactual can be defined by a set of indices where their features differ, i.e., the counterfactual explanation itself:
\begin{equation}
\label{eq:cf_definition}
\delta_{x,c} = \{ i \ | \ x_i \neq c_i \} \\
\end{equation}

\item The counterfactual evidence is irreducible: there's no subset of valid changes that changes the original class.
\begin{equation}
\label{eq:irred}
    \nexists\ \delta'_{x,c} \subset \delta_{x,c}: (\mathcal{M}( \delta'_{x,c}) > t) = (\mathcal{M}( \delta_{x,c}) > t)
\end{equation}
\end{enumerate}

\section{Assessing Feature Importance Values in Counterfactual Explanations}

\subsection{Greedy CFI method}
\label{ssec:greedy}

The greedy CFI method is a simple yet useful approach. It will iteratively select the feature change that makes the largest contribution towards the counterfactual class prediction, as formalized in Algorithm~\ref{alg:calcgreedy}. The resulting scores have some useful properties: the sum of the feature change scores is equal to the difference between the counterfactual and factual prediction ($\mathcal{M}(\textbf{c})-\mathcal{M}(\textbf{\textbf{x}})$); it has a simple interpretation that can easily be presented in charts (further discussed in Section~\ref{sec:greedy}); and it is computationally efficient. Note that Algorithm~\ref{alg:calcgreedy} is not the most efficient implementation of a greedy algorithm, in favor of readability.

This method shares a large similarity with the  SEDC~\cite{martens2011explaining} counterfactual generation approach since both have a similar objective. However, while SEDC is an algorithm designed to find a counterfactual explanation given a factual instance and model, the Greedy CFI method already assumes a counterfactual instance and assigns a score for each feature change. So Greedy CFI is a generic version of the initially proposed `Score Evolution' chart of Martens and Provost. 

\begin{algorithm}
\SetKwFor{RepTimes}{repeat}{times}{end}
\SetKw{Kwin}{in}
\caption{Retrieving the vector with Greedy scores  ($\varrho$) for a factual ($\textbf{x}$) and counterfactual ($\textbf{c}$) instance and prediction model ($\mathcal{M}$). This code is not optimized as vectorization and other approaches can accelerate the calculation and this is done in our package implementation.}
\label{alg:calcgreedy}
\KwIn{Prediction model ($\mathcal{M}$), factual ($\textbf{x}$) and counterfactual ($\textbf{c}$) instance}
\KwOut{Map ($\varrho$) of greedy feature importances for all counterfactual changes in $\delta$}
$\varrho \gets \{\}$\;
$current\_pred \gets \mathcal{M}(\textbf{x})$\;
\RepTimes{$|\delta|$}{
    $best\_pred \gets 0$\;
    $best\_feature \gets -1$\;
    \For{$i$ \Kwin $\delta$}{
        $\textbf{x\_r} \gets \textbf{x}$\;
        $\textbf{x\_r}[i] \gets \textbf{c}[i]$\;
        $pred\_r \gets \mathcal{M}(\textbf{x\_r})$\;
        \If{$pred\_r > best\_pred$}{
            $best\_pred \gets pred\_r$\;
            $best\_feature \gets i$\;
            
        }
    }
    $\varrho[best\_feature] \gets best\_pred - current\_pred$\;
    \tcp{update for next iteration}
    $\textbf{x}[best\_feature] \gets \textbf{c}[best\_feature]$\;
    $current\_pred \gets best\_pred$\;
    $\delta \gets \delta \setminus best\_feature$\;
}
\KwRet{$\varrho$}
\end{algorithm}

\subsection{CounterShapley CFI Method}
Given the non-linearity of the complex models, different sequences may give different importance values for the same feature change. Therefore, we introduce the CounterShapley CFI method next, which presents a way to calculate importance values independent of a specific modification path. As this relies on Shapley values, we will first provide a brief introduction to this game theoretical concept, followed by a short discussion of the previous use of Shapley in machine learning.

\subsubsection{Shapley Values}
\label{sub:svshap}

Shapley values quantify the contribution of different variables to a final numeric result ($\varphi$). This method is derived from game theory, where each variable is considered a player $p$ whose combinations can form coalitions $V$ that contribute to a certain outcome according to a gain function $\mathcal{H}(V)$~\cite{shapley1953quota}. Therefore, we can calculate the contribution ($\varphi_i$) of a certain player $p_i$ in a coalition formed by a set of $|Z|$ players by considering its contribution in all possible coalitions:

\begin{equation}
\label{eq:originalshapley}
    \varphi_i(\mathcal{H}) = \sum_{V \subseteq Z \setminus \{i\}}\frac{|V|!(|Z|-|V|-1)!}{|Z|!}\big(\mathcal{H}(V \cup \{i\})-\mathcal{H}(V)\big)
\end{equation}
The total number of terms Equation \ref{eq:shapley} will have to add together to calculate the Shapley value for a single player equals:
\begin{equation}
\label{eq:n_calc_1_feature}
    N_{coalitions, i} = \sum^{|Z|-1}_{i=0} \binom{|Z|-1}{i}=  2^{|Z|-1}-1
\end{equation}
However, to have a complete calculation of all players' contributions, one must consider every possible subset of players, and not just the ones excluding $i$. This then yields the following number of iterations:
\begin{equation}
    N_{coalitions} = \sum^{|Z|}_{i=0} \binom{|Z|}{i} = 2^{|Z|}-1
\end{equation}
These Shapley values have several desirable properties~\cite{winter2002shapley}:
\begin{enumerate}
    \item Symmtery: two players' contributions are the same, only if they contribute the same to every coalition that does not contain either player.
    \item Efficiency: the sum of all players' contributions is equal to the total worth of the game or the grand coalition.
    \item Dummy: a player with zero contribution does not affect the outcome contribution of a coalition, independent of the coalition in which it occurs.
    \item Linearity: if one takes the sum of two gain functions, then the Shapley value of any player as scored by this sum of gain functions is the same as when one calculates the Shapley value for each gain function, and sums the results. Similarly, when scaling up the gain function by a factor of $a$, this will simply yield $a$ times the Shapley value as calculated by the original gain function.
\end{enumerate}

\subsubsection{Shapley Values in Machine Learning - SHAP}
The assignment of feature importance values for individual instances traces back to 2010 with Strumbelj and Kononenko work~\cite{strumbelj2010efficient} and has gained popularity with LIME in 2016~\cite{lime,zafar2021deterministic}. Since then, multiple algorithms have iterated upon this explanation approach to improve its stability~\cite{visani2022statistical,zafar2019dlime}, where the same instance could yield different importance weights. Among the strategies to circumvent this problem, Shapley value properties demonstrate their potential for solving the stability issue demonstrated by LIME. Given the model's non-linear behavior, features are allowed to contribute differently, depending on which coalition of features they are added to. Moreover, the dummy property assures that features that do not contribute to any coalition have no importance either. Also, the efficiency property allows assigning proportional importance values relative to a certain baseline. What exactly this baseline is, depends on what we consider as the model prediction of an ``empty set of players".

SHAP~\cite{vstrumbelj2014explaining} adapts the Shapley values calculation to the machine learning context. As a main novelty, they developed a strategy of unifying an adapted Shapley value calculation and  feature importance methods. The features are considered to be the players, and the scores obtained by the model are meant to describe, similarly to Shapley's gain function, how each feature contributes to the final prediction score. It has as a baseline the average prediction score. The calculation is performed through a model agnostic approaches (such as KernelSHAP), or model-specific approaches (such as TreeSHAP) to generate more reliable feature importance values. For KernelSHAP, Shapley values are estimated by performing a least squares regression using Lasso normalization. On the other hand, TreeSHAP~\cite{10.1038/s42256-019-0138-9} is data independent: it does not directly need external data since it uses the model's tree nodes information to calculate the features' marginal contributions.

Despite the success of these approximations, they have limitations that substantially impact their application in explaining machine learning models. For TreeSHAP, the obvious implication is its narrow compatibility to tree-based models only. Although KernelSHAP does not have this constraint, its calculation can be computationally expensive for a large number of features and training points: it can take up to $L\cdot2^M$ calculations, where $L$ is the sample number of (training) instances, and $M$ is the number of total features. This creates a trade-off between stability (large sample) and computational efficiency (small sample).
Moreover, even though SHAP yields more stable results over different runs when compared to LIME, its feature importance values can still change depending on the sampling since both the baseline score and the coalitions rely on the randomly selected data.
Finally, the coalitions considered by SHAP can be unlikely or even include impossible data  points. This is due to the fact that SHAP adapts feature values in order to compare instances to a baseline case. For instance, if we consider one-hot encoded (OHE) features, some SHAP coalitions can consider multiple active features from the same OHE (e.g., concurrently being married and single). This issue has broader consequences with KernelSHAP, since it uses multiple data instances that are not necessarily in the same local region as the instance of interest. For this reason, it can end up giving deceptive importance weights to features~\cite{ghalebikesabi2021locality} for a local context.

Finally, SHAP is unable to assign importance scores to counterfactual explanations. It is not compatible with comparing pairs of instances, nor feature \textit{changes}, because it considers the average prediction as a baseline. Despite that, SHAP could still be useful given its variety in plotting resources, filling the gap that counterfactual explanations generally fail to deliver engaging visual resources.

\subsubsection{CounterShapley definition}

As discussed earlier, counterfactual explanations consist of a factual instance to be explained, and a counterfactual point with a different class. Although this type of explanation has benefits, discussed in Section~\ref{sec:motivation}, such as simplicity and sparsity, this representation alone does not precisely show each feature change's influence on this classification change. Consequently, we cannot tell which feature changes contribute more to the score change than others. This directly impacts the quality of the model's explanation since we can only describe the class change and feature modifications. The former is trivially inherent to a counterfactual explanation, while the latter does not directly reflect the impact of each individual feature modification on the model score.

Augmenting the informative value of counterfactual explanations would leverage not just one but multiple explanation methods since counterfactual generation algorithms can have different objectives: SEDC~\cite{martens2014explaining}, for example, tries to improve sparsity and CADEX~\cite{moore2019explaining} minimizes euclidean distance in feature space. Other algorithms target both metrics at the same time, such as DiCE~\cite{mothilal2020explaining} and ALIBI~\cite{looveren2021interpretable}.

As described in Section~\ref{ssec:greedy}, we can assign importance values to the features changes in counterfactual explanations by using their impact over the prediction score. However, the Greedy CFI method considers a specific modification pattern that may not be useful when overall relative importance is desired, especially in highly non-linear models. To solve that issue, we can adopt a strategy similar to SHAP and use the Shapley value approach (introduced in Section~\ref{sub:svshap}) to calculate counterfactual explanation feature importance values.

Let us consider the counterfactual evidence set $\delta_{x,c}$ and associated feature values $x_i$ and $c_i$, such that the factual feature values have to be changed from $x_i$ to $c_i$ in order to go from a factual $\textbf{x}$ to a counterfactual $\textbf{c}$. If we define the model $\mathcal{M}$ as the scoring function, $\delta_{x,c}=\delta$ and $|\delta| = K$ for notational ease, we can use Equation~\ref{eq:originalshapley} to define a Shapley value $\varphi_i$ for the features-to-be-changed $\delta$ as:

\begin{equation}
\labelAndRemember{eq:shapley}{
    \varphi_i(\mathcal{M}) = \sum_{V \subseteq \delta\setminus\{i\}} \frac{|V|! \left(K - |V| - 1\right)!}{K!}\left(\mathcal{M}\left(V \cup \{i\} \right) - \mathcal{M}\left(V \right)\right)}
\end{equation}
where $K$ is the number of features, and the sum extends over all sets $V$ that are a subset of $\delta \setminus \{i\}$. This weighted sum is simply the average marginal contribution of $x_i$ to the class switch, as predicted by the model $\mathcal{M}$. In more detail, from right to left, this equation can be understood by:
\begin{enumerate}
    \item Evaluating the model outcomes of some factual instance $\textbf{x}$, but adapted to have all features changes in $V$, as well as the modification of the feature of interest  ($x_i = c_i$).
    \item Evaluating the same model outcome for the same adapted instance, but without changing the feature of interest.
    \item Considering the difference in these two model outcomes
    \item Weighing this difference, depending on how often this particular coalition of feature changes may occur.
    \item Considering all coalitions of all sizes that do not include a particular feature change, defined by index $i$.
\end{enumerate}

If we want to know how much some feature change contributed to the change in model prediction score, we can change these features from the factual to the counterfactual value one by one, each time evaluating the difference in the model outcome. Given the non-linearity of complex models, the order in which these features are changed may have an influence on the difference in the model outcome. For example, changing some feature value of an instance if two previous values have already been changed can yield a different outcome than changing the exact same feature if no previous changes have been made yet. For this reason, we must consider every possible order in which some feature can be changed, justifying the application of Shapley values. This is a major difference of CounterShapley compared to the Greedy approach, as the latter considers a fixed sequence of modifications.

\subsubsection{Difference between Shapley, SHAP and CounterShapley}
Similarly to SHAP, the CounterShapley approach inherits all the desirable properties of Shapley values. It is fundamental to highlight that, despite the similarity with SHAP, CounterShapley cannot be directly compared with it. They have divergent characteristics and objectives, as shown in Table~\ref{comparison-table}, and explained in detail below. 

\paragraph{Baseline}
The baseline in Shapley is originally just the value of the empty coalition where no players are involved. In contrast, for KernelSHAP, the baseline consists of the average model's prediction over all available data points. For CounterShapley, the baseline is the model prediction of the factual instance that is to be explained. This is a completely different baseline, which is simpler to understand and trivial to calculate.

\paragraph{Coalitions}
The information needed to calculate the importance values is also different for the three methods. Despite their similar goal of calculating the marginal contribution of different coalitions, they require different coalitions. For Shapley values, the coalitions are formed by all possible combinations of players.
For KernelSHAP, the coalitions are formed by all possible feature value combinations of the sampled data points; the contribution values are then obtained by calculating the average prediction score of the instances with the same replaced features for the different points being considered.
CounterShapley's coalitions are all the possible combinations of feature value differences between the factual and counterfactual instance, i.e. $\delta$.

\paragraph{Complexity}
With these considerations on what each method calculates, we can estimate their complexity. All have an exponential complexity, but we observe that CounterShapley does not necessarily consider all features (or players). When a counterfactual algorithm is optimized to minimize the number of features that are to be changed in order to create a counterfactual, this can greatly reduce the complexity of assessing the importance of these features. In Appendix~\ref{appendix:example}, we illustrate how the calculation of CounterShapley works for one feature change, which also explicitly illustrates its complexity. Apart from the fact that KernelSHAP requires more features than CounterShapley, it also has a higher complexity because it requires more data instances. CounterShapley only needs $2$, while KernelSHAP needs a decent number before it can assess the feature importances with some degree of stability.

\paragraph{Efficiency}
Additionally, their efficiency property is also notably different. Shapley values have their sum equal to the value of the scoring function when all players are being considered. When we sum the KernelSHAP values we have the difference between the average prediction of the sample and the explained instance prediction score. For CounterShapley, the sum of their values is equal to the difference between the counterfactual and factual prediction scores.

\paragraph{Goal and applicability}
These different baselines, coalitions, calculations, and sums highlight their distinct interpretations and applicability. CounterShapley considerably differs from SHAP since it aims to explain the feature's effect on the factual to the counterfactual change, while SHAP compares the feature of interest with the average prediction. SHAP is more suitable if one wants to understand how every single feature influences the prediction scoring, and CounterShapley is advisable for assigning importance weights to counterfactual explanations. Additionally, CounterShapley does not create the explanation, but rather augments the informative value of pre-existing counterfactual explanations. Therefore, if the speed of generating an explanation is a relevant matter, one must also include the counterfactual generation time, which can vary greatly depending on the counterfactual generator being applied~\cite{de2021framework}.

\begin{table}
\begin{tabular}{ll}
\rowcolor[HTML]{EFEFEF} 
                        &                                                                  \\
\rowcolor[HTML]{EFEFEF} 
\multicolumn{2}{c}{\cellcolor[HTML]{EFEFEF}\textbf{Shapley}}                               \\
\textbf{Objective}      & Given a gain function $\mathcal{H}$ and $Z$ players that can form $2^{Z}$ coalitions (V), \\
                        &calculate each player`s contribution to all possible coalitions. \\
                        & \\
\textbf{Base value}     & Output value of empty coalition $\mathcal{H}(\emptyset)$.   \\
                        &                                                                  \\
\textbf{Information needed}    & For any coalition $V$, its output value with ($\mathcal{H}(V \cup \{i\})$) and \\
                        & without ($\mathcal{H}(V)$) player's contribution.                         \\
                        &                                                                  \\
\textbf{Calculations}   & $\mathcal{O}(2^Z)$ (all coalitions).                            \\
                        &                                                                  \\
\textbf{Values' sum}    & Gain function's value when all players are considered $\left(\mathcal{H}(all\ players)\right)$.  \\
                        &                                                                  \\
\textbf{Interpretation} & Marginal contribution of any player over the final result considering all players.     \\
\rowcolor[HTML]{EFEFEF} 
                        &                                                                  \\
\rowcolor[HTML]{EFEFEF} 
\multicolumn{2}{c}{\cellcolor[HTML]{EFEFEF}\textbf{KernelSHAP}}                                  \\
\textbf{Objective}      & For a prediction model $\mathcal{M}$ and a (sampled) dataset $\textbf{D}$ with $N$ rows and $M$ columns, \\
                        & calculate each feature importance of a data point $\textbf{x}_e$.  \\
                        & \\
\textbf{Base value}     & Average model's prediction over all data points $\left(\sum_{i=0}^{N} \frac{\mathcal{M}(\textbf{x}_i)}{N}\right)$.   \\
                        &                                                                  \\
\textbf{Information needed}& Model's prediction value for all combinations between target ($\textbf{x}_e$) and \\
                        & data points $\left(\forall \textbf{x}_i \in \textbf{D}_{N,M}\right)$.\\
                        &                                                                  \\
\textbf{Calculations}   & $\mathcal{O}(N2^M)$.          \\
                        &                                                                  \\
\textbf{Values' sum}    & Difference between the average prediction over the sampled data points and                       \\
                        & the instance point to be explained $\left(\mathcal{M}(\textbf{x}_e)-\sum_{i=0}^{N} \frac{\mathcal{M}(\textbf{x}_i)}{N}\right)$.                              \\
                        &                                                                  \\
\textbf{Interpretation} & Average contribution to the model's prediction of each feature in relation \\
                        & to sampled data points. \\
\rowcolor[HTML]{EFEFEF} 
{\color[HTML]{9B9B9B} } & {\color[HTML]{9B9B9B} }                                          \\
\rowcolor[HTML]{EFEFEF} 
\multicolumn{2}{c}{\cellcolor[HTML]{EFEFEF}\textbf{CounterShapley}}                        \\
\textbf{Objective}      & For a factual ($\textbf{x}$) and counterfactual  ($\textbf{c}$) instance with $K$ modified features out of \\
                        &$M$ features in total, calculate the contribution of each feature change\\
                        & according to the model $\mathcal{M}$. \\
                        & \\
\textbf{Base value}     & Factual instance`s prediction score $\left(\mathcal{M}(\textbf{x})\right)$.                                 \\
                        &                                                                  \\
\textbf{Information needed}    & Model's prediction value for all possible combinations of feature                                                                   \\
 & changes ($\delta$) in the factual instance ($\textbf{x}$). \\
                        &                                                                  \\
\textbf{Calculations}   & $\mathcal{O}(2^K)$ $(K \leq M)$.                            \\
                        &                                                                  \\
\textbf{Values' sum}    & Difference between counterfactual and factual predictions $\left(\mathcal{M}(\textbf{c})-\mathcal{M}(\textbf{x})\right)$.  \\
                        &                                                                  \\
\textbf{Interpretation} & Contribution of all counterfactual features and their respective changes  \\  & compared to the factual prediction score.  
\end{tabular}
\caption{Comparison table of Shapley values and two model agnostic approaches for explaining machine learning predictions, highlighting the differences between Shapley, KernelSHAP and CounterShapley.}
\label{comparison-table}
\end{table}

\subsubsection{Implementation of CounterShapley}
We continue by describing how CounterShapley values can be calculated, highlighting the methods' theoretical differences and advantages, and show how it lends itself very nicely to visualisation purposes. Algorithm~\ref{alg:dictgeneration} presents a method of generating a map between any-sized coalitions of feature changes and their corresponding model prediction outcomes. This map can be used in conjunction with Algorithm~\ref{alg:calcshapleyvalue}  to calculate the Shapley value of the change in some feature $i$. Since Algorithm~\ref{alg:dictgeneration} has already concerned itself with mapping all possible coalitions to their corresponding model prediction outcome, Algorithm~\ref{alg:calcshapleyvalue} does not need to perform any further calls to the model's prediction method and can simply fetch them from the result of Algorithm~\ref{alg:dictgeneration}. This is an important aspect of not just this specific implementation, but any implementation, as the algorithmic complexity is defined by the number of calls to this model prediction method. The extension to calculate all Shapley values for all features $\delta$ is now trivial, and can be done by using Algorithm~\ref{alg:calcshapleyvalue} for every change in feature $i$ in $\delta$.

\SetKwFunction{getCombinations}{getCombinations}
\SetKw{Kwin}{in}
\begin{algorithm}
\caption{Generating marginal feature contributions for varying coalition sizes}
\label{alg:dictgeneration}
\SetKwInOut{KwIn}{Input}
\SetKwInOut{KwOut}{Output}
\KwIn{factual ($\textbf{x}$), counterfactual ($\textbf{c}$), model ($\mathcal{M}$)}
\KwOut{$G = \{coalition: score\}$}
$G \gets \{\}$\;
\For{$L \gets 0$ \KwTo $K$}{
    $subsets \gets \getCombinations{V, L}$\;
    \For{$V$ \Kwin $subsets$}{
        \tcp{create adapted instance to score}
        \tcp{start with the factual instance}
        $\textbf{c}^* \gets \textbf{x}$\;
        \tcp{change out feature values}
        $\textbf{c}^*[V] \gets \textbf{c}[V]$\;
        \tcp{save score of adapted instance}
        $G[V] \gets \mathcal{M}(\textbf{c}^*)$\;
    }
}
\KwRet{G}
\end{algorithm}

\begin{algorithm}
\SetKwFunction{subsetlen}{getCoalitionsOfLength}
\caption{Calculating the Shapley value of some feature change with index $i$}
\label{alg:calcshapleyvalue}
\KwIn{Feature $i$, map of all coalitions $V$ and corresponding scores $G$}
\KwOut{CounterShapley value $\varphi_i$ of feature's $i$ respective change}
$\varphi_i \gets 0$\;
\For{$L \gets 1$ \KwTo $K$}{
    $C \gets \subsetlen{L, G}$\;
    \For{$V$ \Kwin $C$}{
        \tcp{construct adapted instances}
        \tcp{one that has feature i}
        $with\_i \gets \textbf{x}$\;
        $with\_i[V] \gets \textbf{c}[V]$\;
        \tcp{and one that does not}
        $no\_i \gets \textbf{x}$\;
        $no\_i[V \setminus \{i\}] \gets \textbf{c}[V \setminus \{i\}]$\;
        \tcp{marginal contribution}
        $diff \gets G[with\_i] - G[no\_i]$\; 
        \tcp{add normalised marginal contribution}
        $\varphi_i \gets \varphi_i + diff \cdot L! \cdot (K-L-1)! / K!$ \;
    }
}
\KwRet{$\varphi_i$}
\end{algorithm}

\section{Visualizing Counterfactual Explanations}\label{sec:visualizingCFE}

In this section, we present three different chart types that increase the informative value of counterfactual explanations. Two of them, Greedy and CounterShapley charts, use the CFI methods described in the previous section. The third chart type, the Constellation chart, uses the same calculations as CounterShapley since it analyzes all possible coalitions with the counterfactual feature modifications.

\subsection{Greedy Chart}
\label{sec:greedy}

The greedy chart aims to illustrate the sequential nature of the changes made by the greedy CFI method. We argue that this method brings value to graphical representations for two main reasons. First, as said previously, the calculation of scores has lower computational costs if compared to CounterShapley, from an exponential $O(2^K)$ to a lower complexity $O(K^2)$ behavior. Second, this approach gives us a fixed path that can be depicted in charts, governed by a greedy selection of feature changes. The latter can be useful if the receiver of the counterfactual explanation wants to increase, as much as possible in each modification step, the prediction score. Moreover, this greedy CFI method gives the same importance scores as CounterShapley if the model has a linear association between features. Thus, for lower-complexity models, it can give a sufficiently good approximation of the CFI values obtained with the CounterShapley method. To exemplify its use, let us consider the following counterfactual explanation:

\begin{center}
\begin{tcolorbox}[enhanced,width=10cm,center upper,drop fuzzy shadow southwest,
    boxrule=0.4pt,sharp corners,colframe=yellow!80!black,colback=yellow!10]
\textbf{Explanation 2: Explanation for a loan application decision}
\noindent\rule{\textwidth}{0.4pt}
If your age were 30 instead of 20, salary were \$2,200 instead of \$1,500, sex were F instead of M, and state were CA instead of NY, the application would be classified as approved instead of rejected.
\end{tcolorbox}
\end{center}

\begin{figure}[h]
\centering
\includegraphics[width=14cm]{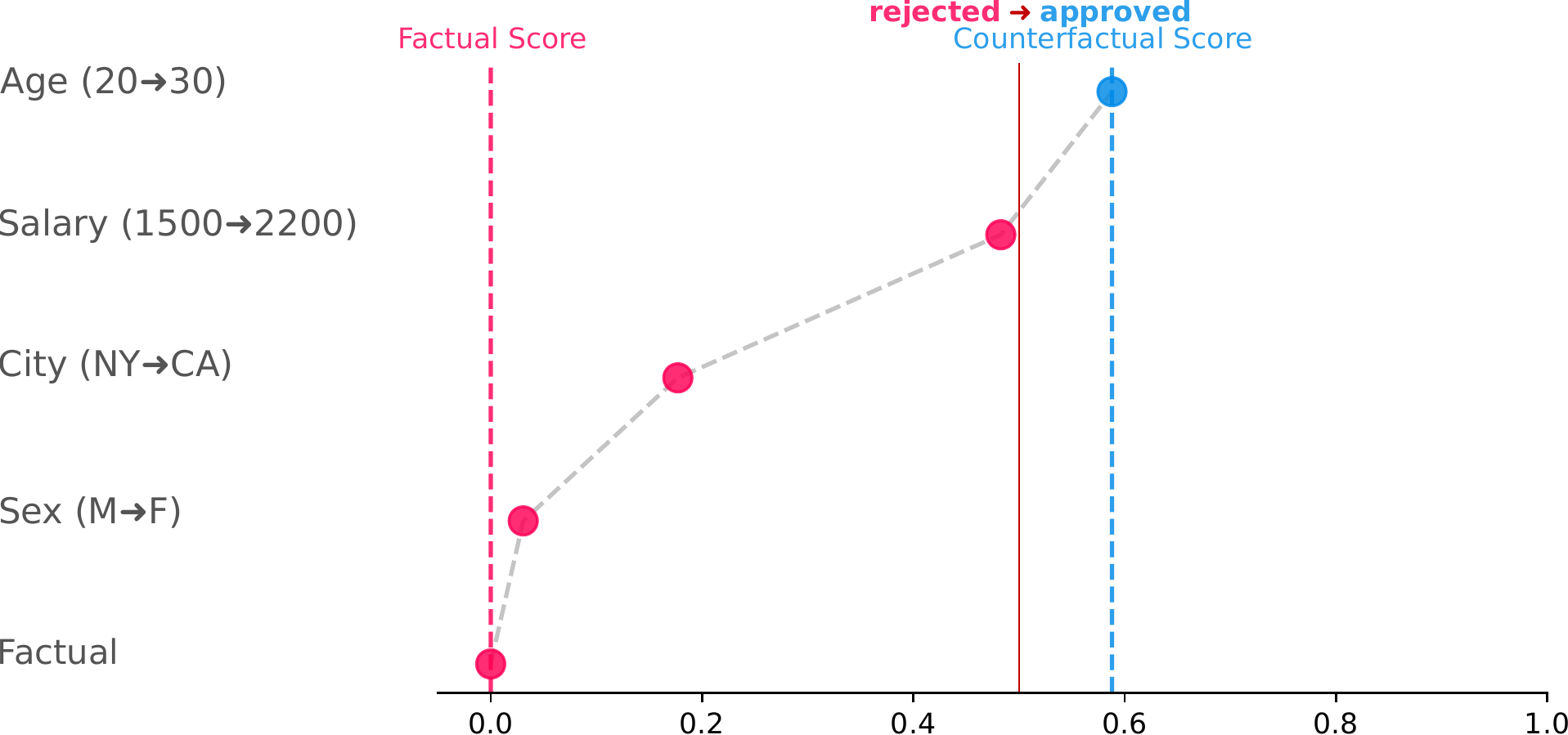}
\caption{Counterfactual explanation visualization for \textit{Explanation 2} using the Greedy chart.}
\label{fig:examplegreedy}
\end{figure}

Figure~\ref{fig:examplegreedy} depicts how \textit{Explanation 2} would be displayed using the Greedy chart. The first point, at the bottom, represents the prediction score of the factual point, which has the features \textit{Sex}, \textit{City}, \textit{Salary}, and \textit{Age} equal to ``M'', ``NY'', $1,500$, and $20$ respectively. The next point (with a prediction score of about $0.35$) shows the change in prediction when the feature \textit{Sex} is changed from ``M'' to ``F''. This modification (if compared to the three others) is the one that leads the highest increase towards the counterfactual class. The next points follow the same pattern, greedily, showing the best feature change to modify the classification result (from rejected to approved). The chart then has a sequence of changes starting from the factual point (bottom left) and gradually selects (from the bottom to the top) the changes that most positively contribute to the counterfactual class, which is depicted on the right side of the red threshold line. This simple chart allows for the visualization of sequential changes and their impact on the prediction score.

It's fundamental to emphasize that this chart represents scores influences in a specific setting: a greedy search. This may not reflect the CounterShapley values in non-linear models. Note for example how changing the City from ``NY" to ``CA" has a larger impact on the model score than changing Sex from ``M" to ``F", but is not the first change in line. This is because this change would not be as influential if the ``Sex" wasn't changed first. Likewise for changing ``Salary": without changing ``Sex" and ``City" first, the impact of this feature change would not be as big as it is. Such is the nonlinear nature of the model. Therefore, practitioners must be aware of whether this chart meets their expectations in explaining a decision, and possibly avoid it if they want to provide a precise estimation of feature change impact over the prediction score of complex models.

\subsection{CounterShapley Chart}

The importance scores of counterfactual feature changes allow us to create a new visualization, here referred to as a CounterShapley chart. Our main objective with this chart is to represent counterfactual explanations in a simple but informative way which allows easy user understanding of what features lead to a class change and how important each of those features is.

\begin{figure}[h]
\hspace*{-1.5cm}  % in a two-column format, this graph should span the entire page
\includegraphics[width=20cm]{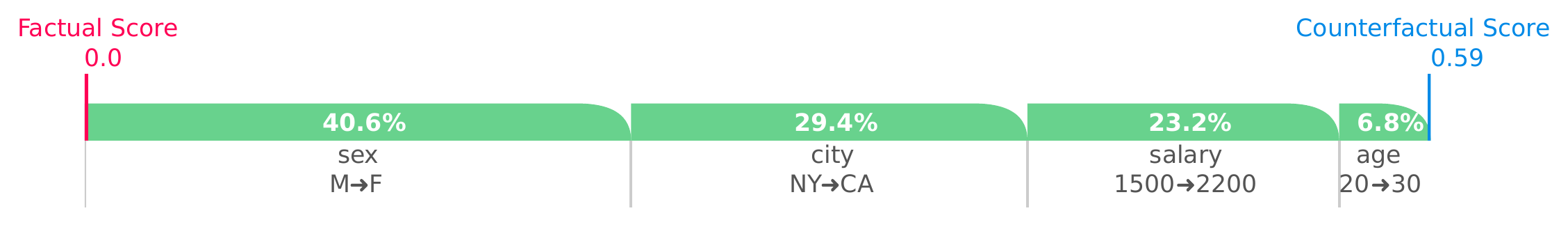}
\caption{Counterfactual explanation visualization for \textit{Explanation 2} using CounterShapley chart. It highlights the factual and counterfactual scores and, in the middle, the contribution to this prediction modification for each feature change in the counterfactual explanation.}
\label{fig:exampleiris}
\end{figure}

Figure~\ref{fig:exampleiris} shows the CounterShapley chart for the counterfactual \textit{Explanation 2}, summarizing diverse information in a single image. The bars represent the CounterShapley values for each feature, and inside them we indicate their contribution in percentage to the total CounterShapley value. Below the bars, we have the feature names and their respective value changes. In addition, the chart shows the original instance, counterfactual instance, and the model prediction score after performing all feature changes. Although all this information could be described in a textual counterfactual, as mentioned earlier, the graphical representation is beneficial~\cite{ainsworth2003effects} to understanding and interpretation.

It's interesting to note how the information conveyed here differs from Figure~\ref{fig:examplegreedy}. The Greedy chart tells us that ``Sex" is the first best change one can make, despite the fact it does not increase the score by a lot initially. The CounterShapley chart tells us that, on average ``Sex" has the biggest effect on the prediction score when averaged over every possible combination of features. The Greedy chart shows that ``Salary" has the biggest impact, but only if ``Sex" and ``City" were changed beforehand. The CounterShapley chart tells us that, when considering every possible combination of feature changes, ``Salary" only accounts for $23.2\%$ of the overall change in prediction score. Thus, the large impact of ``Salary" in the Greedy chart is only because the previous feature changes allowed it to make as big of an impact. Indeed, the CounterShapley value of ``Sex" is large ($40.6\%$), indicating that ``Salary" impacts the prediction score a lot on the Greedy chart, but only because ``Sex" is included in the set of feature changes. ``Sex" does not have a large contribution on its own, but it allows other features to make a big impact.

\subsection{Constellation Chart}

Counterfactual explanations tend to have sparse feature changes, as we discussed in previous sections. The Constellation chart makes good use of this characteristic to display how each feature change combination influences the counterfactual prediction score. This chart has the same complexity as calculating the CounterShapley values, since it iterates over all feature combinations ($2^K$). Contrarily to both previous charts, it also has a more complex representation.

However, this type of chart can be useful to practitioners to have a complete view of how each counterfactual feature combination affects prediction scoring, allowing an extensive analysis of feature changes' influences. Moreover, it can also serve as a debugging tool for counterfactual explanations, since it is possible to verify if any subset of changes crosses the decision threshold, essentially finding a subset of the counterfactual that's also a valid counterfactual (see property 3 of the definition of counterfactuals in Section~\ref{sec:counterfactual_explanations}).

\begin{figure}[h]
\centering
\includegraphics[width=16.5cm]{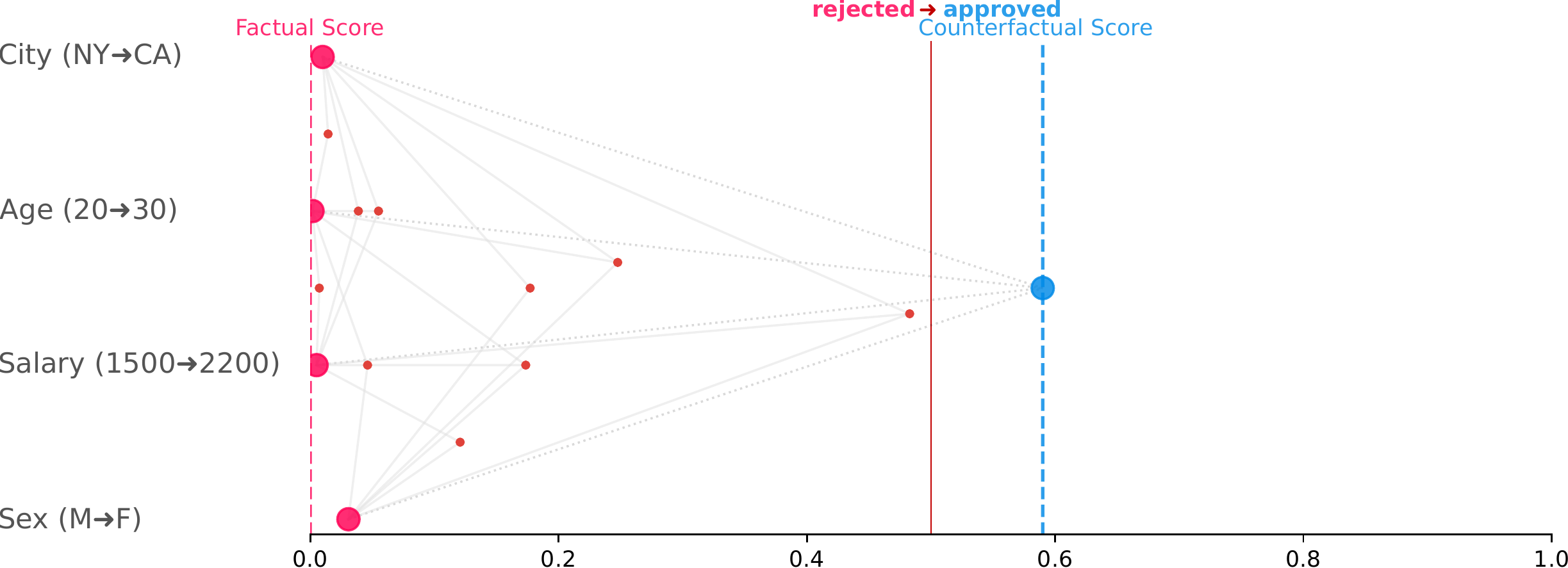}
\caption{Counterfactual explanation visualization for \textit{Explanation 2} using the Constallation chart.}
\label{fig:exampleconstellation}
\end{figure}

It has a similar application in describing importance values to counterfactual changes as the CounterShapley chart, but instead of summarizing them in single values, it detangles all possible feature's change association impacts over the model's prediction score. In Figure~\ref{fig:exampleconstellation} we take the same \textit{Explanation 2} counterfactual to generate a Constellation chart.

The Constellation chart representation in Figure~\ref{fig:exampleconstellation} shows how each single feature change (``Age'', ``Salary'', ``City'',  and ``Sex'') and their combinations affect the prediction score. On the left, we have the four counterfactual changes listed. The first pink dashed line (on the left) shows the factual score according to the $x$-axis. Next to the right, the four larger pink dots represent a single feature change. For example, the bottom large pink dot indicates the prediction score (according to the $x$-axis) when the feature ``Sex" is changed from ``M'' to ``F''. Similarly, the other large pink dots correspond to the feature changes at the same height. The smaller pink dots represent the combinations of different feature changes. For instance, the first small pink dot (from the bottom to the top) indicates the prediction score when both features ``Sex" and ``Salary" are modified (this can also be deducted from the lines connecting to the larger dots). Notice that the height of the combination dots is equal to the average height of the features combined, but this height has no explicit meaning other than visualisation purposes. The rightmost point represents the prediction score when all feature changes are applied, as indicated by the lines connecting to all large dots on the left.

While this view looks more complex than the other previous methods, it also targets a more specialized public that is interested in understanding the underlying meaning of CounterShapley values and features' effects. For example, this chart already shows how the exclusion of ``Age" still yields an instance that's very close to the decision threshold. Note for example how: ``Sex" is indeed the first best feature one can change (lower left big pink dot), conform with the Greedy chart; and, on average, most small pink dots that are linked with ``Sex" have indeed higher prediction scores than those not linked with ``Sex", explaining its large CounterShapley value as seen in the CounterShapley chart. Finally, all small pink dots that are connected to both ``City" and ``Sex" yield the highest values. This is also congruent with the fact that these have the biggest CounterShapley values, as seen in the CounterShapley chart.

Moreover, this chart can help in other technical tasks, such as model improvement, since the multiple feature associations can reveal unexpected effects that may require further investigation. This will be shown in the next section.

\section{Empirical Experiments}
In the previous sections, we have introduced two main algorithms to assess the feature importance values of counterfactual explanations: CounterShapley and a Greedy approach. These counterfactual feature importance values now make it possible to directly compare counterfactual explanations with other feature importance methods, such as LIME~\cite{lime} and SHAP~\cite{vstrumbelj2014explaining}; but also to white-box models, such as the weights of a logistic regression.

In the following sections, we generate various counterfactual explanations using NICE~\cite{brughmans2021nice}, but the results work equally well for other counterfactual methods such as DiCE~\cite{mothilal2020explaining}, ALIBIC~\cite{looveren2021interpretable}, or SEDC~\cite{fernandez2020explaining}. The CFI values are compared to those of SHAP and LIME for various test cases. We show different experiments with multiple data, models, and methods that characterize how our CFI methods work, highlighting the differences with LIME and SHAP. In the first two sections, we use artificial data to demonstrate the characteristics of our methods, while the other three subsequent sections use real-world data, showing its applicability. We also explain cases in which the plots proposed here embed critical value for enhanced explanations and ``debugging'' counterfactual explanations. All experiments presented here are fully reproducible and can be found in the ``Experiments'' branch of our GitHub repository (see below).

Note that these comparisons are not quality assessments of these methods, as there is no consensus on what a ``correct'' explanation should be~\cite{vilone2021notions}. However, empirical testing is useful in providing a better hint at what each explanation approach tries to achieve.

Our work uses Python to make all calculations regarding the counterfactual scores, and the Matplotlib package~\cite{Hunter:2007} to create the graphical elements. Our implementation is open-source and independent from a specific algorithm, which makes it possible to integrate it with any counterfactual generation algorithm. The repository can be found on GitHub \href{https://github.com/ADMAntwerp/CounterPlots}{https://github.com/ADMAntwerp/CounterPlots}. All empirical analyses use Python and open-source packages such as Scikit-Learn~\cite{scikit-learn}, Pandas~\cite{reback2020pandas}, and Seaborn~\cite{Waskom2021}. 

\subsection{Simple Model Experiments}
We use three simple, explainable (white-box) machine learning models (logistic regression, decision tree, and K-nearest neighbors) together with artificially generated data, and explain the predictions with three importance attribution methods: LIME, SHAP, and CounterShapley. Since we need a counterfactual generator for the CounterShapley calculation, we opt for NICE~\cite{brughmans2021nice} to create the counterfactuals. We can then infer if the results given by explanation methods follow an expected pattern or not, by using the intrinsic explainability of these simpler models: feature weights of the logistic regression, decision nodes for the decision tree, and the overall data distribution for KNN.

\begin{figure}[h]
\centering
\includegraphics[width=14.5cm]{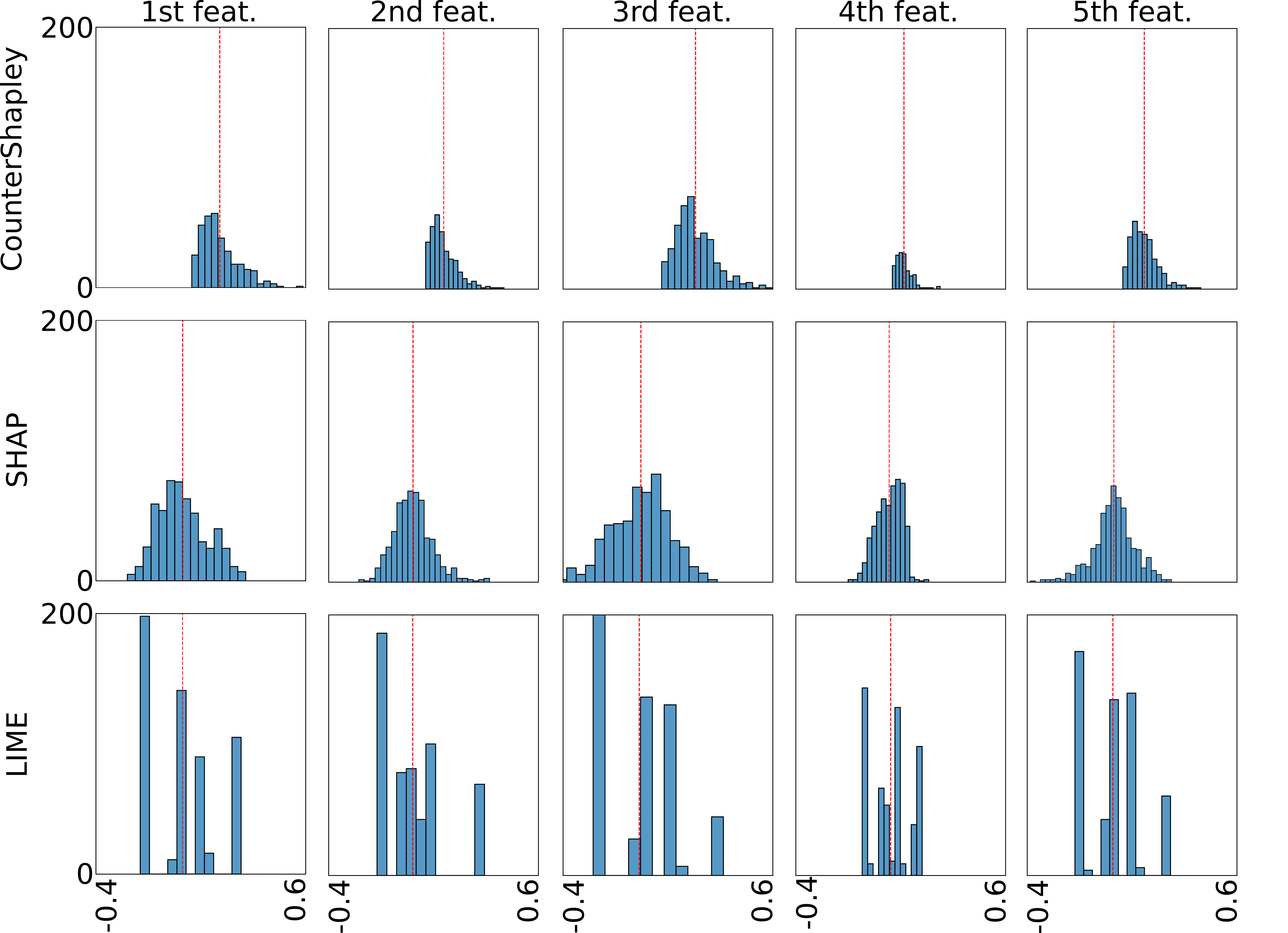}
\caption{Histogram with importance scores of all points that were classified as 0 in the logistic regression classification model, each column represents one feature while each row represents the CounterShapley, SHAP, and LIME distributions, respectively. The red dotted line indicates the distribution's average.}
\label{fig:filrhist}
\end{figure}

For the logistic regression model, we explain 1,000 instances from a 5-feature synthetically generated dataset with two classes and a fixed random state (42) created with Scikit-Learn. Table~\ref{logisticregressionweights} shows the logistic regression weights for each feature. Since individual explanations may not provide enough information to compare each method, we analyze the CFIs by aggregating the explanations that contain non-zero values (Table~\ref{lrfeatureimportances}) and by plotting them in histogram charts (Figure~\ref{fig:filrhist}).

\begin{table}[h]
\centering
\caption{Logistic regression weights for the dataset's experiment features.}
\label{logisticregressionweights}
\begin{tabular}{lrrrrr}
\toprule
Features & 1st &       2nd &       3rd&        4th &        5th \\
\midrule
LR Weight& -0.59 & 0.67 & 0.78 & -0.34 & -0.59 \\
\bottomrule
\end{tabular}
\end{table}

\begin{table}[h]
\centering
\caption{Share, in percentage, of the explanations with non-zero importance score for the logistic regression classification model.}
\label{lrfeatureimportances}
\begin{tabular}{lrrrrr}
\toprule
Features &     1st &     2nd &     3rd &     4th &     5th \\
\midrule
CounterShapley &  58.29 &  53.30 &  75.04 &  25.49 &  53.83 \\
SHAP        & 100.0 & 100.0 & 100.0 & 100.0 & 100.0 \\
LIME        & 100.0 & 100.0 & 100.0 & 100.0 & 100.0 \\
\bottomrule
\end{tabular}
\end{table}

The first main observation we can take from Figure~\ref{fig:filrhist} is the fact that the CounterShapley CFI method generally yields only positive importance scores, while the other methods include negative scores. Negative scores simply represent that an increase in these features leads to a probability decrease that an instance belongs to class 1. It highlights the substantially divergent interpretation from the CounterShapley values, which measure the impact of the feature changes on the class probability, knowing that the sum of all changes should be positive. A negative feature value here would mean that, not only is there at least one coalition including this feature that \textit{decreases} the probability of a class switch, but these also outweigh the overall effect of the remaining coalitions that \textit{increase} the class probability. Either there are more coalitions that have a negative effect on the class probability; or the negative effects are, on average, stronger than the positive effects. As long as the coalition including all feature changes (i.e. the counterfactual itself) yields a class flip, such feature changes can occur in counterfactual explanations. Cases like these are rare, but not impossible.

Second, although the average importance scores for SHAP and LIME are similar, their distribution is considerably divergent. SHAP appears to favor small values, as seen by the peak around $0$, while LIME has an undefined distribution, generally assigning high-importance values to features,

Finally, let's compare the logistic regression weights (in Table~\ref{logisticregressionweights}) to the share of explanations that each feature is included (with an importance score different than zero). We see that the feature with the highest weight (3rd) is the most frequent in CounterShapley values; on the other hand, the other methods always assign scores for all features. While, again, this does not represent an advantage, it shows that CounterShapley is sparser given that it is based on counterfactual explanations.

\begin{figure}
\centering
\includegraphics[width=12cm]{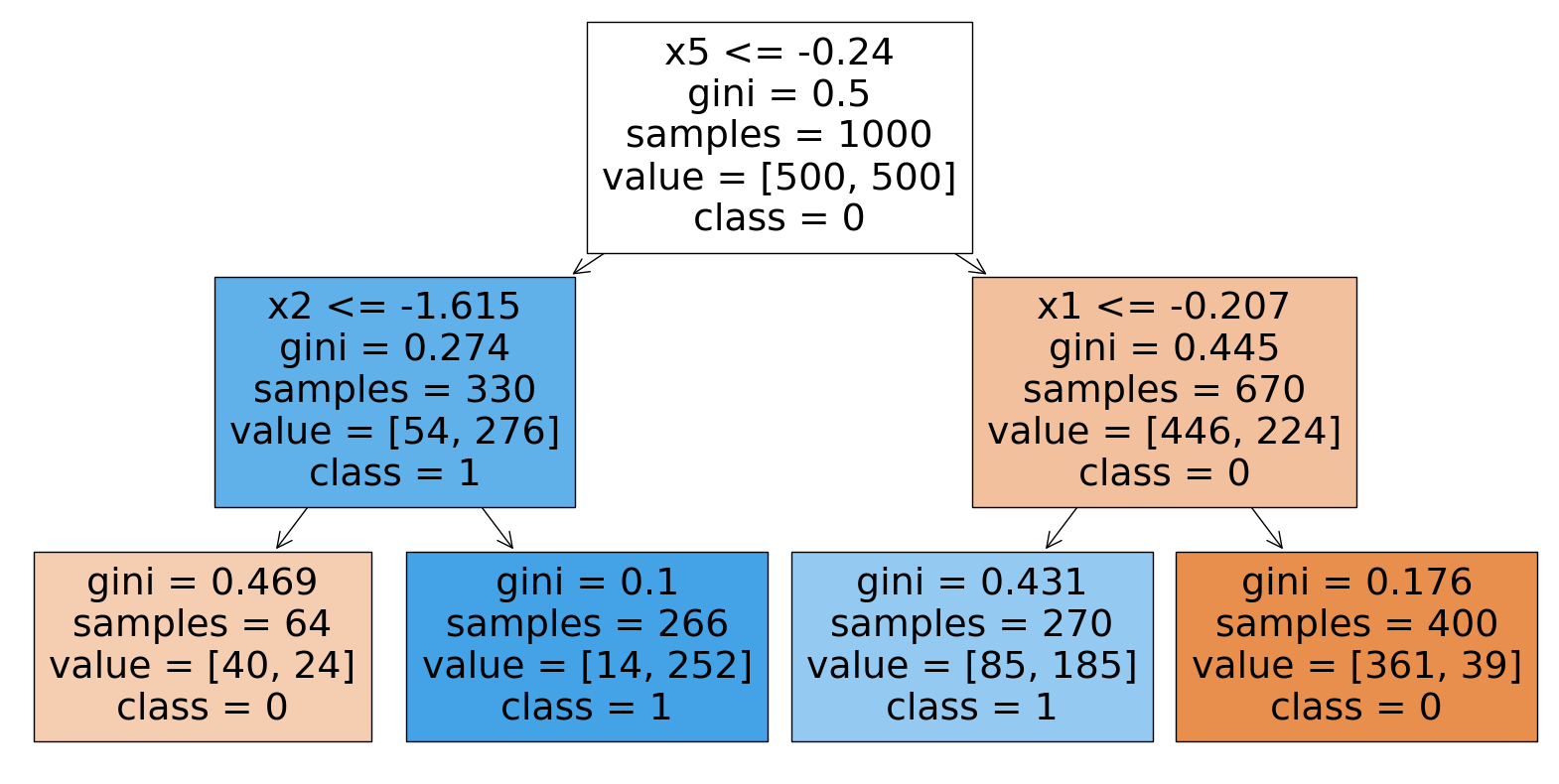}
\caption{Simple decision tree model used for classification, in the first row of each decision node we show the condition, following, it presents the gini coefficient, number of total samples, the share of instances in each class (0 and 1, respectively) and, finally, the major class. For the node leaves, there's no condition statement and the majority class is the model's prediction class.}
\label{fig:dtempirical}
\centering
\includegraphics[width=14.5cm]{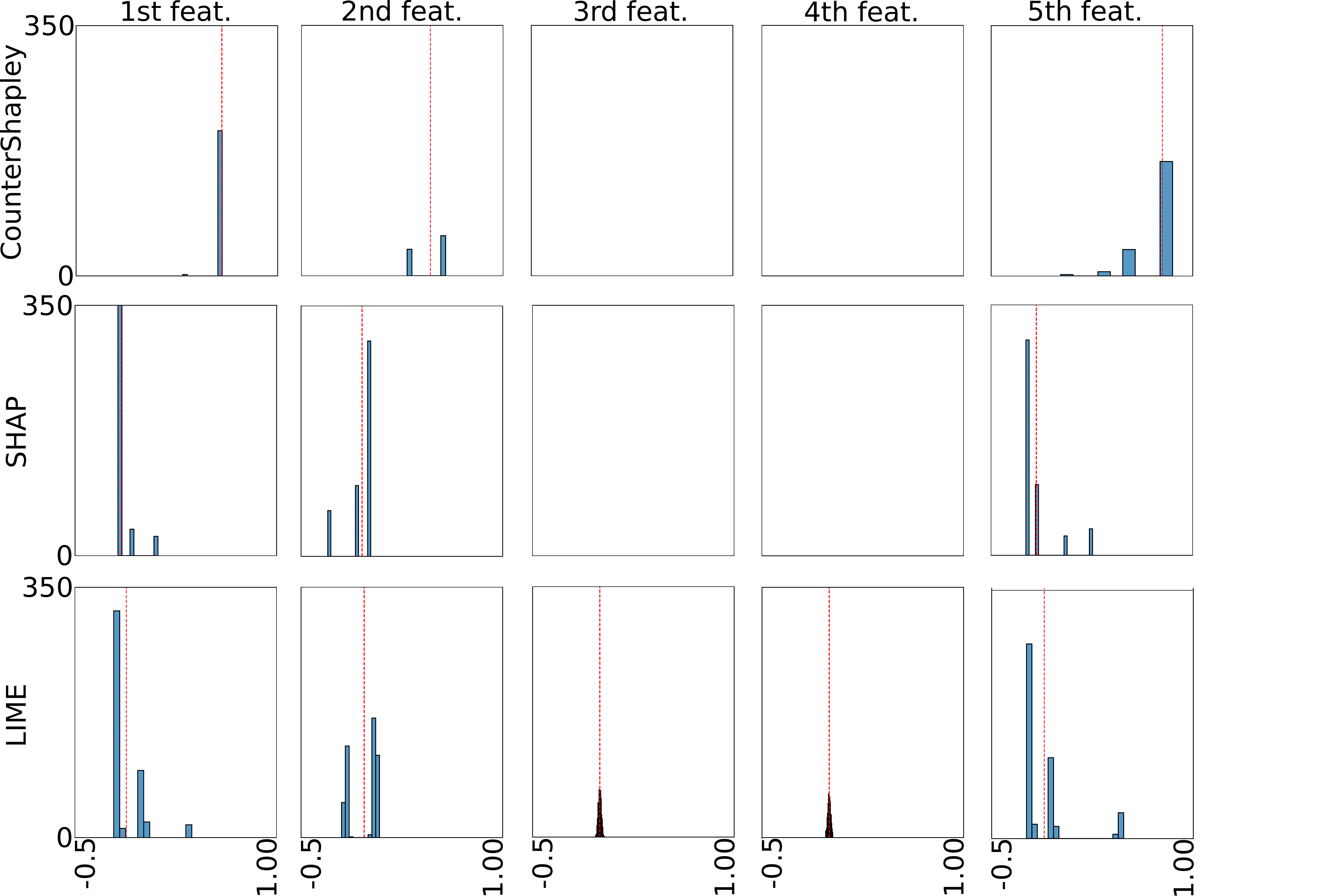}
\caption{Histogram with importance scores of all points that were classified as 0 in the decision tree classification model, each column represents one feature while each row represents the CounterShapley, SHAP, and LIME distributions, respectively. The red dotted line indicates the distribution's average. For the 3rd and 4th features in the CounterShapley and SHAP methods, the histograms are empty because all importance scores are equal to zero.}
\label{fig:fidthist}
\end{figure}

For the decision tree, we made a very simple model with a depth of 2 and a total of 3 decision nodes using the same artificially generated data that we used for the logistic regression experiments. Figure~\ref{fig:dtempirical} shows a representation of the model, where we can observe that only 3 (out of the 5 features) are present in the model: the first, second, and fifth. Figure~\ref{fig:fidthist} then shows the distribution of the importance scores obtained by the feature scoring methods.

\begin{table}[h]
\centering
\caption{Share, in percentage, of explanations instances in which importance scores are different from zero for the decision tree model.}
\label{dtfeatureimportance}
\begin{tabular}{lrrrrr}
\toprule
Features &         1st &         2nd &         3rd &         4th &         5th \\
\midrule
CounterShapley &  44.18 &  20.00 &  0.0 &    0.0 &  44.18 \\
SHAP        & 100.0 & 100.0 &  0.0 &  0.0 & 100.0 \\
LIME        & 100.0 & 100.0 & 100.0 & 100.0 & 100.0 \\
\bottomrule
\end{tabular}
\end{table}

\newpage

Again, we see a similar trend regarding the negative values assigned to importance scores for SHAP and LIME, which, similarly, means that increasing their respective features decreases the probability of class 1. Moreover, including the analysis of Table~\ref{dtfeatureimportance}, we see that the 2nd and 3rd features have 0 importance for CounterShapley and SHAP, which is understandable (and expected) since these two features are not present in the decision tree nodes. Although LIME assigns importance scores to the 2nd and 3rd features, they are relatively low values compared to the other scores. Regarding the absolute importance values, it is clear on CounterShapley gives high importance to the 5th feature since it is the first splitting node and for the 1st feature since it splits more instances than the 2nd feature.

\begin{figure}[h]
\centering
\includegraphics[width=8cm]{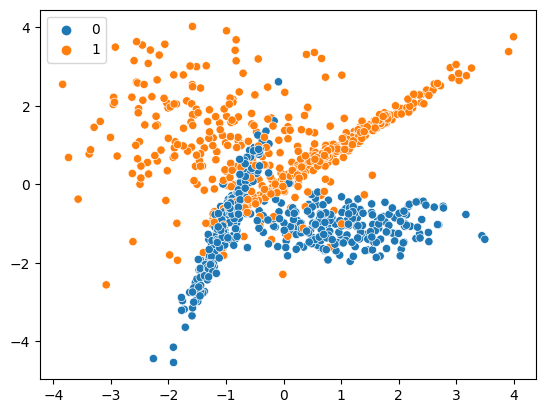}
\caption{Instances classes of the dataset used to the KNN classifier. Blue points have class 0, while the orange points are classified as 1.}
\label{fig:knndataset}
\end{figure}

\begin{figure}[h]
\centering
\includegraphics[width=7cm]{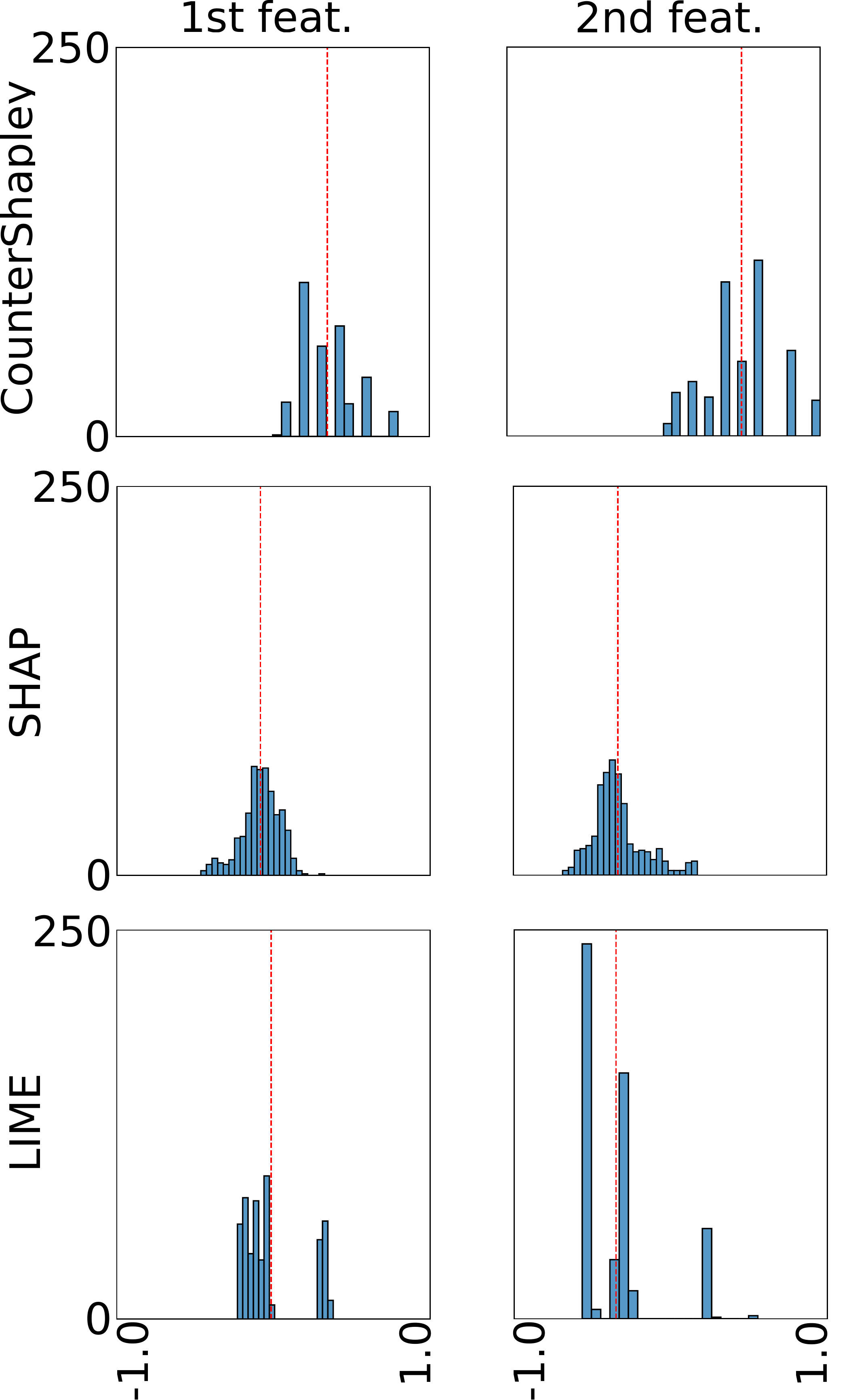}
\caption{Histograms with importance score distribution of 0 classified instances in the KNN model for the 1st and 2nd features (columns) and different methods (rows).}
\label{fig:fiknnhist}
\end{figure}

Finally, the last explainable model was a KNN classifier fitted in a 2-features dataset depicted in Figure~\ref{fig:knndataset}. This dataset is also artificially created with Scikit-Learn, with two informative features and classes, a fixed random state (equal to 42) to guarantee reproducibility, and all other parameters as default. The results of the average importance scores are shown in Figure~\ref{fig:fiknnhist}.

\begin{table}[h]
\centering
\caption{Share of importance scores different from 0 of the KNN classifier instances using three different methods.}
\label{knnfeatureimportance}
\begin{tabular}{lrr}
\toprule
Features &     1st (x) &     2nd (y) \\
\midrule
CounterShapley &  62.26 &  81.61 \\
SHAP        & 100.0 & 100.0 \\
LIME        & 100.0 & 100.0 \\
\bottomrule
\end{tabular}
\end{table}

All methods assign a higher (absolute) importance to the 2nd ($y$-axis) feature. This is understandable since, by analyzing the data distribution in Figure~\ref{fig:knndataset}, we can observe that the $y$-axis makes better segregation between classes than the $x$-axis. Similarly to the previous cases, in Table~\ref{knnfeatureimportance}, we see again the sparsity of the CounterShapley explanations and also that the most important feature (the 2nd) is more frequent in the explanations.

Considering all three experiments with the simple models described above, we see that CounterShapley assigns importance scores that are substantially different from SHAP and LIME. One major difference observed in all experiments is the fact that CounterShapley values focus on the score changes needed to flip the classification - given the already mentioned decision-driven nature. This characteristic can make the score interpretation easier since both LIME and SHAP assign importance scores to every class (in the case of binary classification, both 0 and 1 classes). However, we highlight that, depending on the research question, multiple importance scores for classes may be a desirable outcome, with the practitioner being on a task to consider each case in particular. The experiments also present, in general, the intrinsic model's justifications (weights, decision nodes, and data distribution) correspond to the average model's explanations. Consequently, all methods are aligned with the main features responsible for the classification. Nevertheless, as thoughtfully discussed in the previous sections,  the numerical scoring values and intrinsic meaning of CounterShapley values are not the same as SHAP and LIME.

\subsection{Simple Data Experiments}

For the next experiments, we evaluate how CounterShapley, SHAP, and LIME assign importance scores to feature changes in a complex black box model based on random forest classifiers. However, if we use equally complex data, we wouldn't be able to say if the numerical results correspond to what is expected from the model and data. Therefore, we use simple datasets whose patterns can be clearly grasped by analyzing their class distributions. These data have multiple patterns and follow a similar concept as the experiments performed by Robnik-Sikonja \textit{et al.}~\cite{robnik2008explaining}.

\begin{figure}[h]
\centering
\includegraphics[width=14cm]{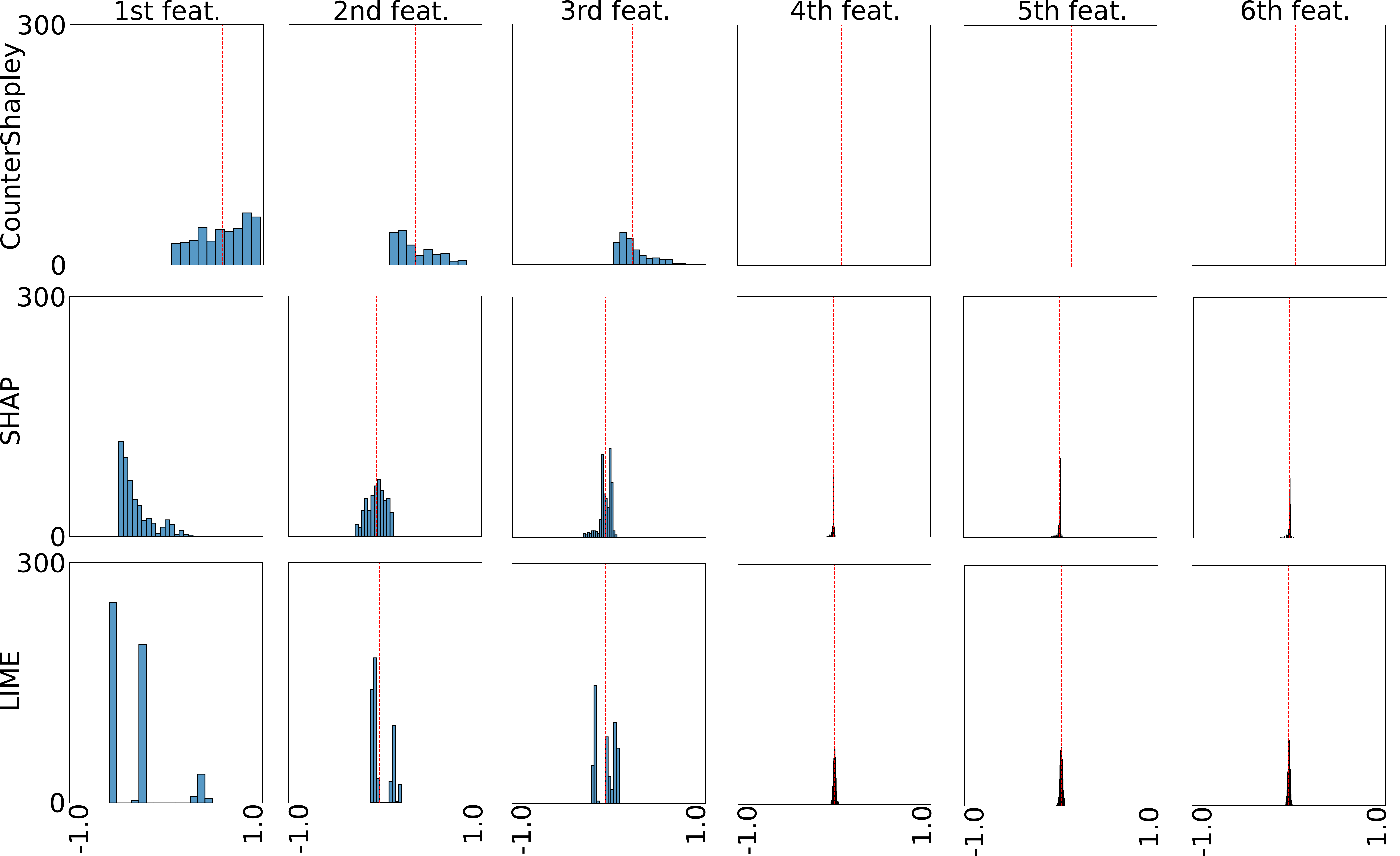}
\caption{Histograms with importance score distribution of 0 classified instances for the three correlated features (1st, 2nd, and 3rd) and the randomly generated features (4th, 5th, and 6th).}
\label{fig:fisimpledatahist}
\end{figure}

Our first experiment evaluates the impact of adding noisy training data by making a very simple dataset consisting of 3 informative features and 3 random features with no direct connection to the class labels. This dataset was artificially generated using the Scikit-Learn package, for a classification task with two labels and a fixed random state (42).  Figure~\ref{fig:fisimpledatahist} shows the distribution of importance score values for informative features (1st, 2nd, and 3rd) and random features (4th, 5th, and 6th). Additionally, Table~\ref{tablenoise} shows the percentage of features that had non-zero importance scores.

\begin{table}[h]
\centering
\caption{Share, in percentage, of explanation instances which includes non-zero importance scores for the three correlated (1st, 2nd, 3rd, highlighted in bold) and three random features (4th, 5th, 6th, unrelated to the label class).}
\label{tablenoise}
\begin{tabular}{lrrrrrr}
\toprule
{Features} &      \textbf{1st} &      \textbf{2nd} &      \textbf{3rd} &      4th &      5th &      6th \\
\midrule
CounterShapley &  83.83 &  35.53 &  31.34 &  4.99 &  6.9 &  3.19 \\
SHAP        & 100.0 & 100.0 & 100.0 & 100.0 & 100.0 & 100.0 \\
LIME        & 100.0 & 100.0 & 100.0 &  100.0 & 100.0 & 100.0 \\
\bottomrule
\end{tabular}
\end{table}

The importance scores show that all three methods assign more relative importance to the informative features if compared to the random ones. For this specific dataset, all three methods also show a similar pattern having the 1st feature as the most important, followed by the 2nd and 3rd. One may question why the random features are not equal to zero, especially because in a previous experiment, in which the model only used a subset of features, both CounterShapley and SHAP assigned exactly 0 importance scores to unused features. The reason is that, differently from the latter case, the random features in this experiment influence the prediction scoring. We also see the characteristics mentioned in the previous experiments, like CounterShapley having positive scores, sparsity with the most important features being more frequent, and average scores different from SHAP and LIME, still happening in this case.

\begin{figure}
\centering
\includegraphics[width=15cm]{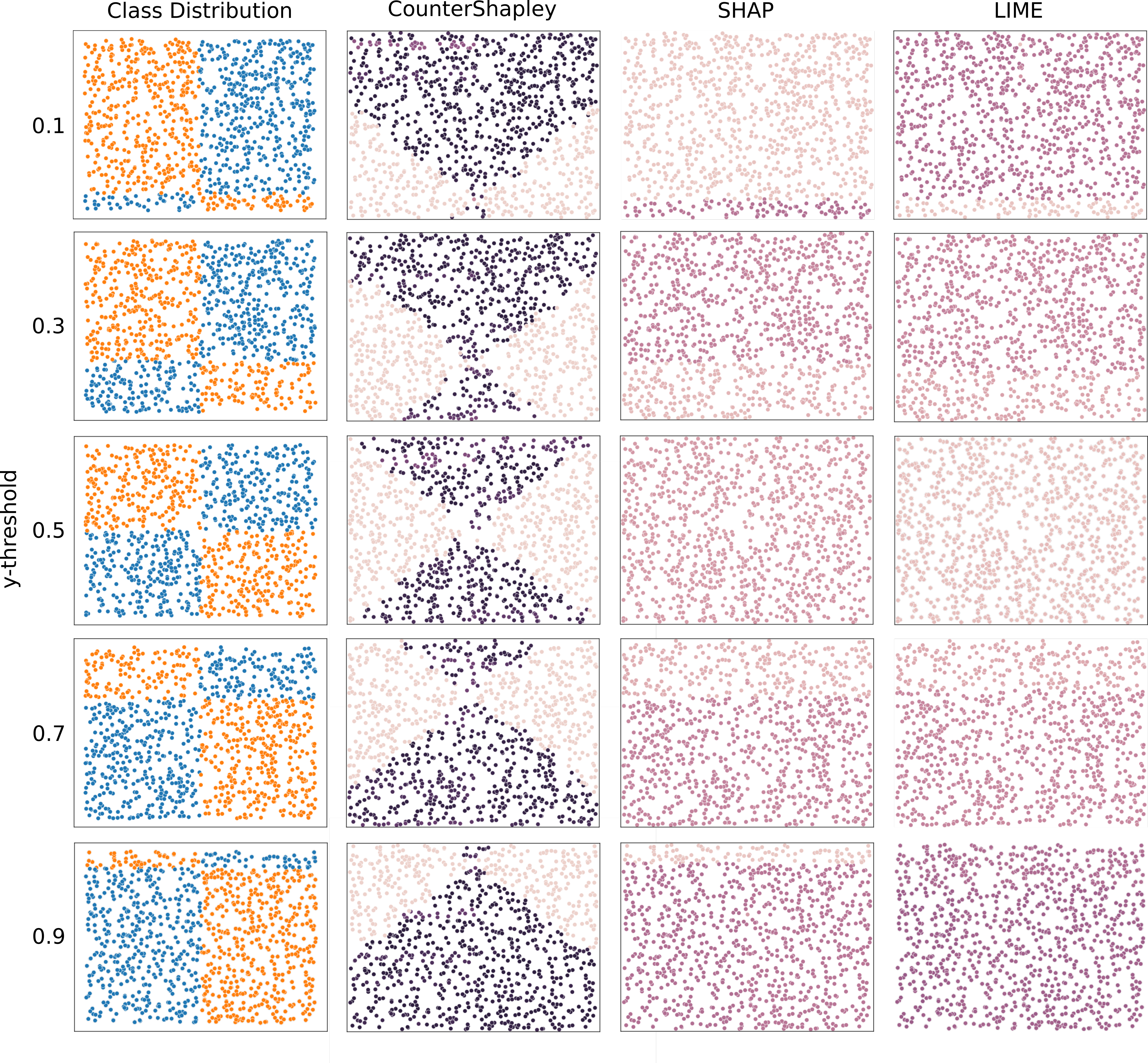}
\caption{First column shows the class distribution (0-class blue, 1-class orange) of each instance in the dataset. The following columns show in a color scale (darker means higher values) the $x$ importance for each instance using CounterShapley, SHAP, and LIME.}
\label{fig:squarefeatureimportance}
\end{figure}

In the following simple data experiments, we use only 2 features, and classes synthetically generated with Scikit-Learn to follow a certain geometric pattern that can be detected by simple visual analysis. The first experiment divides the data points into four quadrants. The separation between the upper and lower quadrants has different thresholds, as shown in Figure~\ref{fig:squarefeatureimportance}. This figure also shows the $x$-axis importance score using CounterShapley, SHAP, and LIME methods. Since we use a binary class and the random forest classifier assigns classes with a high degree of confidence, the $y$-axis importance charts are very similar but with inverse scores, with clear regions being darker and vice-versa.

The charts in Figure~\ref{fig:squarefeatureimportance} show that SHAP and LIME have very similar behavior scoring importance; for the $y$ threshold equal to 0.5, they score about the same score for all points in any region. This can be justified since we have equally distributed classes over the 4 quadrants: therefore, overall, the $x$ and $y$ axis contribute equally to the output prediction. When the threshold is changed, the upper and lower quadrants have different scores for the $x$ feature. This is also a reasonable result because the region with increased height has a lower importance for $y$ since a variation in that direction has a reduced effect on probability while the $x$-axis is unchanged. However, the pattern observed for CounterShapley is substantially different if compared to the other two methods. As previously discussed, counterfactual explanations show the minimal feature change to modify the model's classification. Therefore, with a score based on this concept, the explanations are divided into regions where the $x$ feature is closer to the decision threshold (hereafter having a higher importance score) and regions in which the $y$ feature is closer to the decision threshold (consequently, $x$ has a lower importance score). So for the same data and model, CounterShapley gives different scores if compared to SHAP and LIME. All methods have reasonable justifications for their scoring strategy. However, we emphasize that CounterShapley seems more reasonable if we want to score counterfactual explanations since it uses the same concepts.

\newpage

\begin{figure}[h]
\centering
\includegraphics[width=12cm]{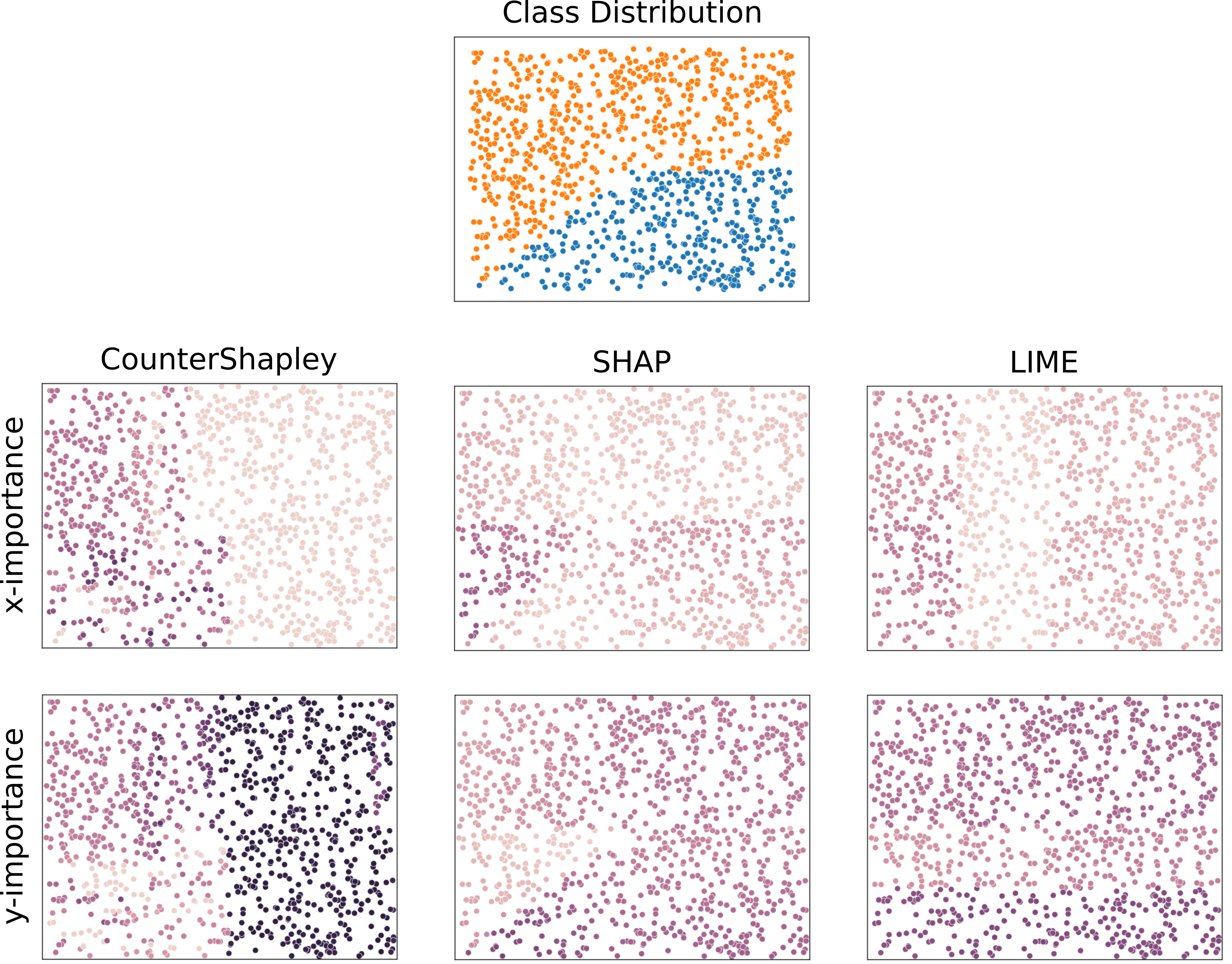}
\caption{Top chart shows the data distribution with class 0 (blue) and class 1 (orange) instances. The second and third row charts show importance weights for the $x$ and $y$ features, respectively.}
\label{fig:linearconstantfeatimportance}
\end{figure}

For the next experiment, the dataset has a pattern with two behaviors. For features up to $x=0.5$ the classification has a linear ($x=y$) threshold, while for values higher than 0.5, it has a constant behavior ($y=0.5$). Figure~\ref{fig:linearconstantfeatimportance} illustrates this dataset class distribution and the importance scores for the $x$ (top) and $y$ (bottom) features. There we can see the complementary effect of importance scores, given that we have a binary classification and a predictor that assigns scores with a high degree of confidence.

The CounterShapley CFI method clearly separates the two regions in the chart, for the linear behavior region ($x<0.5$) does not show a clear preference for either feature while, for the constant region, the $y$ feature has evident higher importance values. This behavior is justified because, in the linear region, the $x$ or $y$ feature distance to the classification threshold is about the same for most instances, while in the constant region, $y$ is usually the closest path to the model's decision threshold. Unlike the previous cases, SHAP and LIME had substantially different patterns; however, both generally give more importance to the $y$ feature. Nevertheless, their importance has a clearly more subtle relationship with the dataset class distribution if compared to CounterShapley importance scores.

\begin{figure}[h]
\centering
\includegraphics[width=12cm]{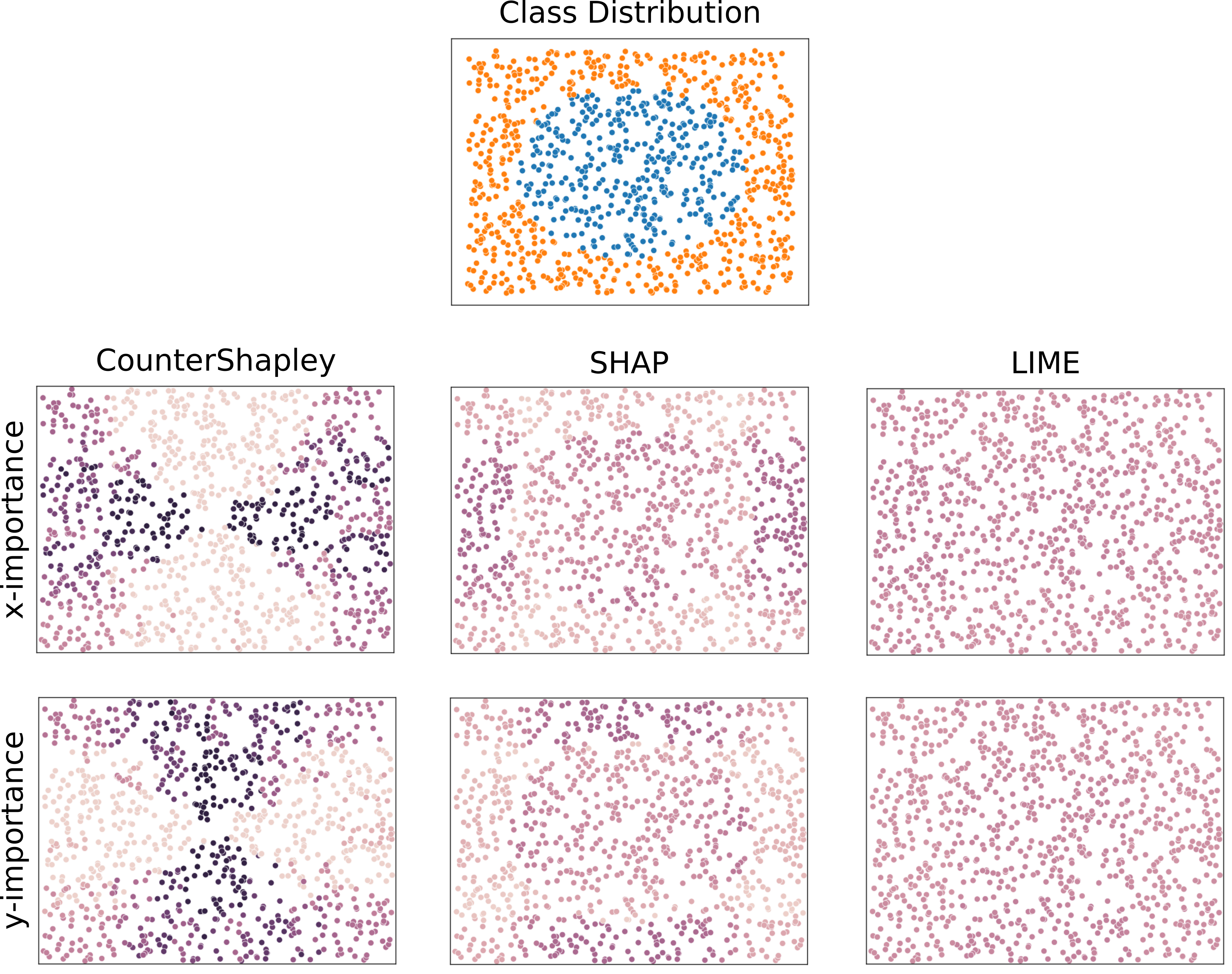}
\caption{Top chart shows the data distribution with class 0 (blue) and class 1 (orange) instances. The second and third row charts show importance weights for the $x$ and $y$ features, respectively.}
\label{fig:circularfeatimportance}
\end{figure}

Figure~\ref{fig:circularfeatimportance} shows the importance scores for a circular pattern in the dataset. In this case, the 0 class points are concentrated in the center, while the 1 class is around the circular region formed by the 0 class instances.

The analysis of importance values shows that LIME assigns about the same importance for the $x$ and $y$ features in all instances. This may happen given the sampling strategy, which randomly selects nearby points which end up normalizing the value given the circular geometry. Interestingly, SHAP and CounterShapley have some similarities where the middle top and bottom regions assign a higher relevance to the $y$ feature while the lateral regions assign more importance to the $x$ feature. However, the internal region of SHAP does not look to have this pattern. The justification for such a pattern in CounterShapley may be explained by the sparsity-optimized strategy that the counterfactual generator uses. Therefore, in lateral regions, the most optimal feature change to reach the decision threshold is modifying the $x$ feature, while it is $y$ otherwise.

All experiments with simple data above show again that CounterShapley values give considerably different importance values for feature changes, following a scoring pattern more consistent with the idea behind counterfactual explanations, minimal sparse changes to change prediction classification. Moreover, we can notice the capacity that CounterShapley values have to react differently to multiple patterns in the same dataset, where these regions are less evident if we use SHAP and LIME. This characteristic is intrinsically linked to the fact that counterfactual explanations are based on only two points (factual and counterfactual). In contrast, SHAP and LIME have a normalizing effect since they consider multiple points over the dataset. This difference does not necessarily represents an advantage of CounterShapley values because all methods present reasonable justifications for their scores. Therefore, practitioners must consider all these characteristics and their objectives before selecting an importance scoring method.

\subsection{Replication Experiments}
\label{sec:replication}

One of the main advantages of the methods presented in this paper is their broad compatibility with any counterfactual generation algorithm. The compatibility is not only in terms of theoretical requirements since they only need the factual, counterfactual, and model prediction functions, but also given our algorithmic framework that can be easily implemented using Python and popular data science packages. In this section, we show how our methods can be applied to three different counterfactual generation algorithms: NICE~\cite{brughmans2021nice}, DiCE~\cite{mothilal2020explaining}, and ALIBIC~\cite{looveren2021interpretable}. We provide a snippet code of how they are implemented, highlighting the simplicity, and we perform some experiments with data and models which they use in their respective papers. All these experiments are publicly available on GitHub in the Experiments branch.

NICE~\cite{brughmans2021nice} is a fast and reliable counterfactual generation algorithm that finds optimal explanations. It uses dataset labeled instances to find a first optimal instance with a different class and then perform multiple optimization iterations to improve the counterfactual with respect to a specific optimization objective (which can be sparsity, proximity, or plausibility). Appendix Algorithm~\ref{niceexample} shows how easily the NICE official package\footnote{\hyperlink{https://github.com/DBrughmans/NICE}{https://github.com/DBrughmans/NICE}} can be adapted to generate all analyses proposed by the methods presented in this article.

\begin{figure}[h]
\centering
\includegraphics[width=12cm]{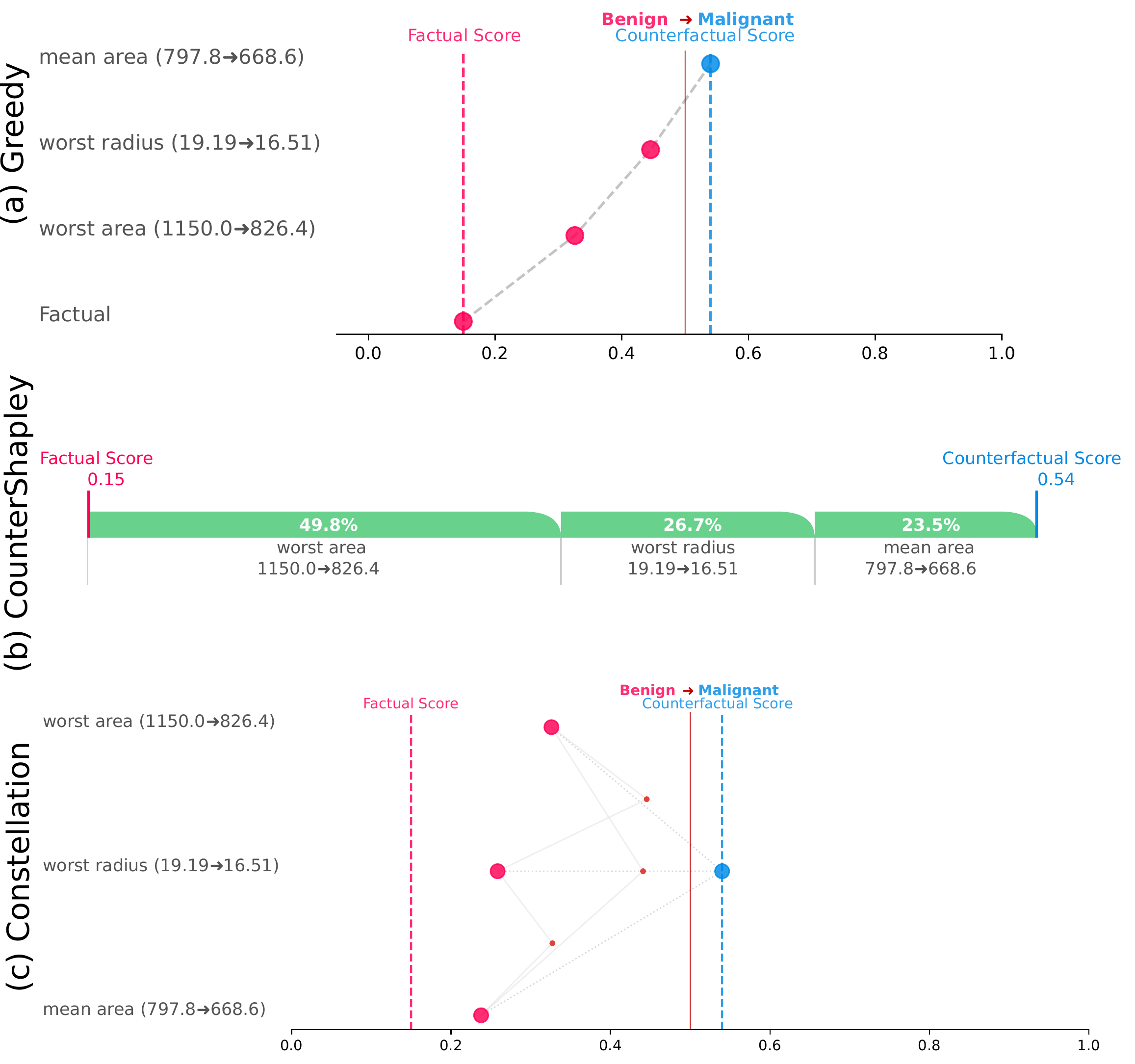}
\caption{Greedy, CounterShapley, and Constellation charts generated using a NICE counterfactual explanation from the Wisconsin Brest Cancer dataset.}
\label{fig:niceexample}
\end{figure}

Our method allows the creation of an object (showed in the Appendix Algorithm~\ref{niceexample}) that works exactly like NICE, since it is an extended class, but we change the output of the ``explain'' function to not being a simple list of the counterfactual points. Rather, the function's output returns ``CounterPlot'' objects, which are able to give the CounterShapley values for each feature and plot any of the three charts presented here (Greedy, CounterShapley, and Constellation). Figure~\ref{fig:niceexample} shows an example of the result obtained from an instance of the Wisconsin Breast Cancer dataset, as used in the NICE paper.

We also use the DiCE's~\cite{mothilal2020explaining} counterfactual generator official Python package\footnote{\hyperlink{https://github.com/interpretml/DiCE}{https://github.com/interpretml/DiCE}}. This counterfactual generator can work independently of a specific model (model agnostic) and allows to generate a set of diverse counterfactual explanations. Appendix Algorithm~\ref{diceexample} shows how a function can take the objects generated by the DiCE counterfactual generation and transform them into the object that allows all analysis presented in this paper.

\begin{figure}[h]
\centering
\includegraphics[width=12cm]{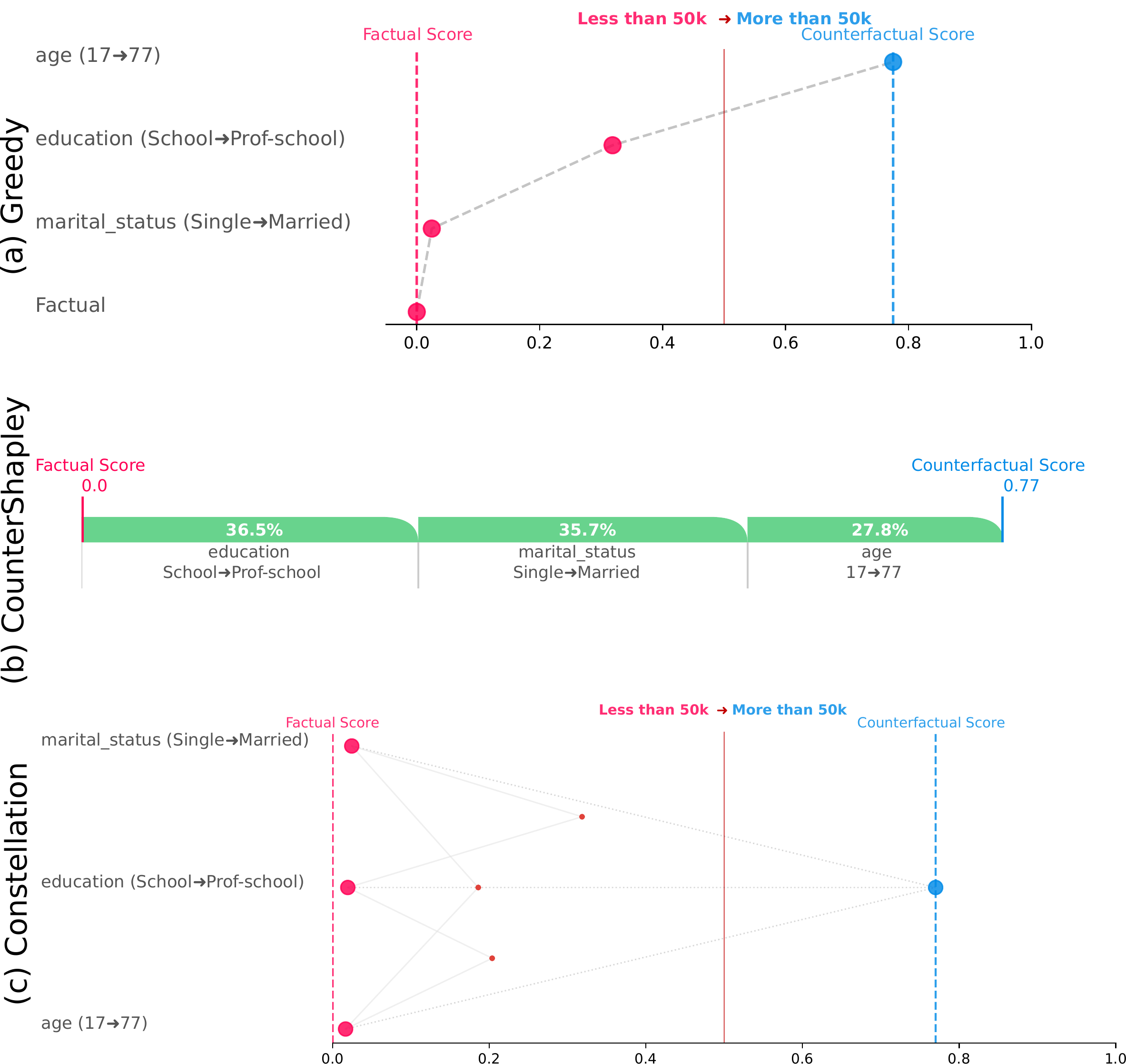}
\caption{Feature change importance charts (Greedy, CounterShapley, and Constellation) for the Adult dataset using DiCE counterfactual generator in a sample instance.}
\label{fig:diceexample}
\end{figure}

Note that, in DiCE's case, the algorithmic framework is more complex since the package is responsible for the dataset encoding, and the resulting counterfactual is inside a custom object. Nevertheless, our algorithm can still be adapted and processes all the diverse counterfactual instances produced by DiCE. As a sample example, we use the same dataset DiCE uses in their package official documentation, the Adult dataset, and Figure~\ref{fig:diceexample} shows the results. In this case, we evaluate the features needed to change the prediction from low-income (class 0) to high-income (class 1).

ALIBIC~\cite{looveren2021interpretable} is also adapted using our framework using Appendix Algorithm~\ref{alibicexample}. This counterfactual generator performs a gradient descent using a complex loss function that includes terms for distance, sparsity, and similarity to the dataset manifold. Similarly to what we have done with DiCE, we use the package official documentation\footnote{\hyperlink{https://docs.seldon.io/projects/alibi/en/stable/examples/cfproto\_housing.html}{https://docs.seldon.io/projects/alibi/en/stable/examples/cfproto\_housing.html}} to get a dataset, generate a model and counterfactual explanation. In this case, we use the California Housing dataset and an artificial neural network as the machine learning model. ALIBIC generates an object that includes multiple attributes, such as the counterfactual point, classes, and features' ranges.

\begin{figure}[h]
\centering
\includegraphics[width=12cm]{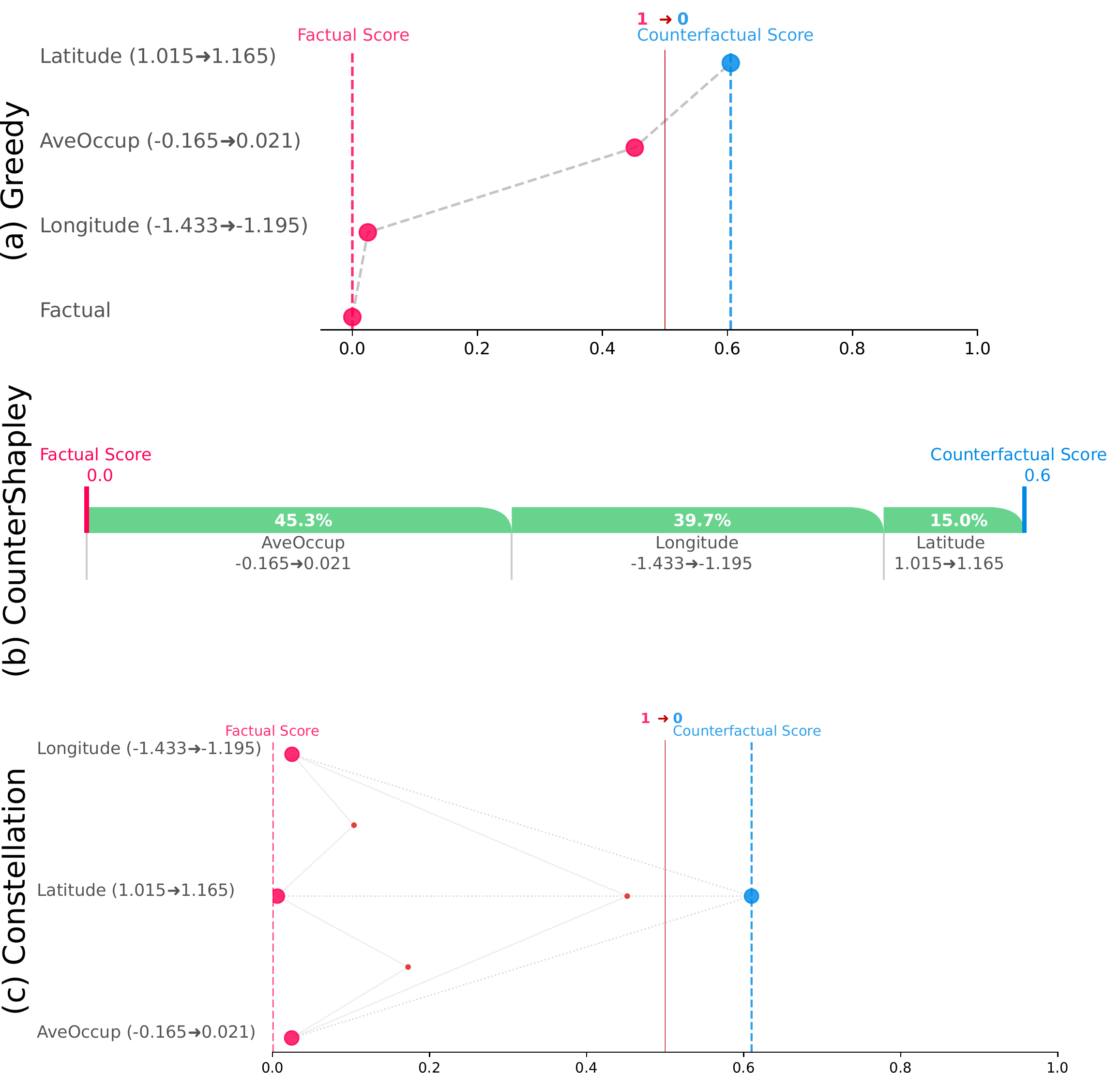}
\caption{ALIBIC feature change importance charts, Greedy, CounterShapley, and Constellation using an instance from the California housing dataset.}
\label{fig:alibicexample}
\end{figure}

ALIBIC's adaptation to generate our analysis is considerably simpler than that of DiCE, as DiCE uses multiple built-in methods from their framework, such as the dataset treatment and model adapter. For ALIBIC (and NICE), we use simpler strategies that directly use the factual point and the prediction model. Figure~\ref{fig:alibicexample} shows an instance explanation example with the three chart types.

With the examples shown above, we showcase that not only the theoretical concepts described in this article are independent of a specific counterfactual generator, but also our algorithmic framework works with any type of generator. Therefore, this represents a substantial contribution to the counterfactual explanation area since it can readily embed a higher explanation value to novel and past algorithms.

\subsection{Missclassification Experiments}

One of the most valuable uses of explanation methods, especially those that focus on local/instance-level explanations, is the investigation of why a model assigned a wrong classification to a given instance. The explanation can reveal patterns or features that can be further investigated, leading to correction or improvements.

Counterfactual explanations, as already mentioned, have multiple characteristics (simple, sparse, decision-oriented, etc.) that make them a valuable method for investigating wrong classification instances. However, the isonomic view of feature changes' importance to the model's prediction may not give enough information for a complete understanding of the underlying causes of the misclassification. Therefore, this experiment shows an example of how the analysis presented in this article can help to unveil more information about the model scoring and decision. To illustrate this application, we use the Bankruptcy Prediction dataset~\cite{liang2016financial}, which includes bankruptcy data from the Taiwan Economic Journal from 1999 to 2009. As a model, we use the Scikit-Learn random forest algorithm with default parameters and a fixed random seed (42).

\begin{figure}[h]
\centering
\includegraphics[width=12cm]{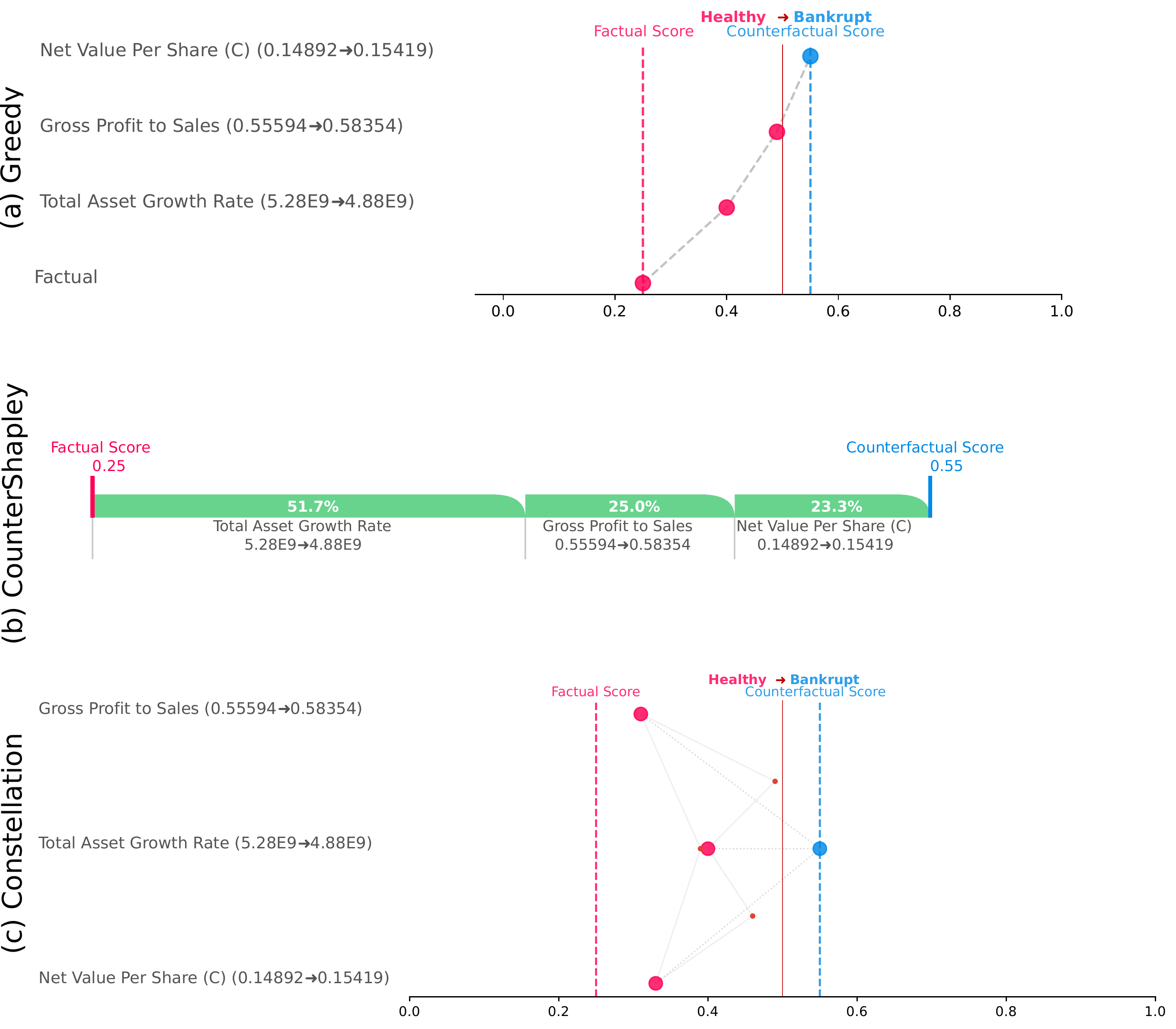}
\caption{NICE counterfactual explanation of an instance from the Bankruptcy Prediction dataset, which was incorrectly classified by the model as a healthy company (but it got bankrupt instead).}
\label{fig:examplewrong}
\end{figure}

\begin{figure}[h]
\centering
\includegraphics[width=11cm]{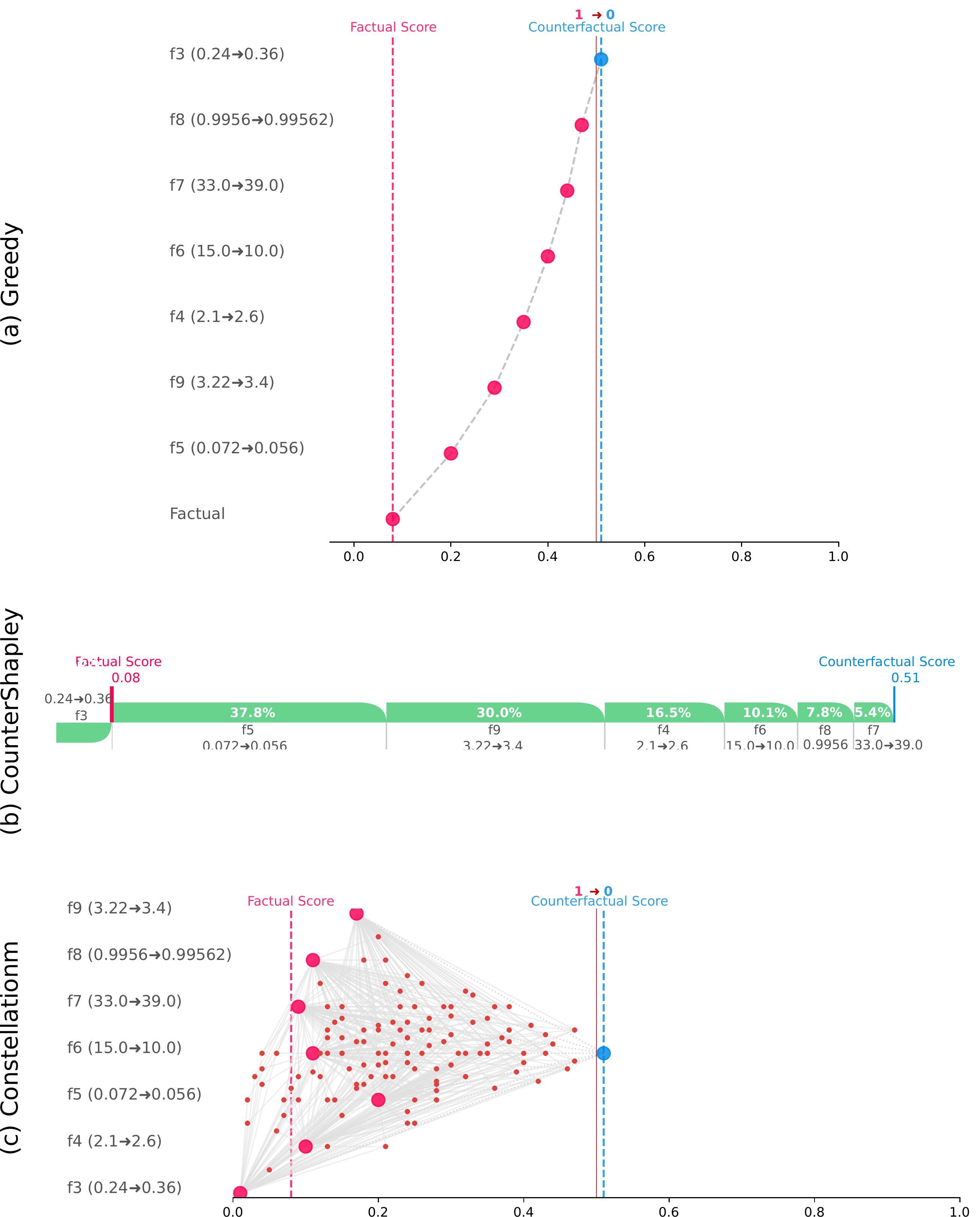}
\caption{Analysis made over a counterfactual explanation generated by NICE that has negative CounterShapley value. Although, for the Greedy chart, this behavior cannot be directly observed, in CounterShapley and Constellation chart is possible to see that \textit{f3} (and some subsets containing it) generate prediction scores lower than the factual score.}
\label{fig:negativesubset}
\end{figure}

Figure~\ref{fig:examplewrong} shows the feature change importance for an instance that is incorrectly classified as a healthy company but got bankrupt. The counterfactual explanation and instances can show interesting insights about this misclassification, such as: a lower total asset growth rate would positively contribute towards a correct classification. This makes sense since slower growth generally is seen as a pessimistic indicator of a company's prospects. As this was the most important feature change, contributing to more than 50\% of CounterShapley value, measures such as the acquisition of more data related to asset growth and feature engineering could potentially help the model achieve better performance. Moreover, the two other features also give us an interesting perspective on the misclassified instance. The second most important feature change (according to CounterShapley values) is a higher gross profit to sales share, and the third, with a similar importance value, is a higher net value per share. At first sight, these changes look unexpected since both rises are associated with positive outcomes. However, it's important to notice these two changes alone are not sufficient to flip the classification and must be associated with the first, most important feature change. This analysis enhances the informative value of counterfactual explanations, allowing us to investigate the pattern formed by these feature changes: maybe the association of higher shares and profit with lower asset growth indicates a purely speculative/momentaneous gain which is not sustainable in the long run. These unexpected associations can also mean that the model is not behaving as wanted, making spurious associations caused by overfitting or another problem in the dataset. In summary, the counterfactual explanations associated with the CFI methods and charts give us extra knowledge of how changes affect the prediction score and classification result, which can lead to better model understanding and improvements.

\subsection{Special Cases}

Finally, in this last section of the experiments, we show two special cases in which the analysis presented in this article could contribute to a better understanding of the model and counterfactual result evaluation. First, we talk about counterfactual feature changes which have negative CounterShapley values, revealing non-linear behaviors between features. Then, we present a way to identify counterfactual explanations that have more feature modifications than needed.

\subsubsection{Negative Contributions}

Complex models can lead to highly non-linear behaviors in which features that usually have a certain behavior (for example, increase the probability of a class) completely change when a specific set of features are present. Simple counterfactual explanations cannot detect such behavior since they only disclose the feature modifications needed to make a change in the classification prediction. However, as shown in Figure~\ref{fig:negativesubset}, our analysis can detect features that, although they have a negative contribution over the prediction scoring, are necessary for a class change.

Our analysis uses the UCI's Wine Quality dataset\footnote{Dataset page: \href{https://archive.ics.uci.edu/ml/datasets/wine+quality}{https://archive.ics.uci.edu/ml/datasets/wine+quality}} and a random forest classifier built with Scikit-Learn and default parameters. In Figure~\ref{fig:negativesubset}, we show one instance example that has a feature (\textit{f3}) with a negative CounterShapley value as seen in its respective CounterShapley chart. This can be further understood with the Constellation chart, which shows that \textit{f3} decreases the inverse class probability if changed alone (bottom clear blue dot) or in association with other features. However, as can also be seen in the Greedy and Constellation chart, this feature is fundamental for the class change since, if absent, it is not sufficient to modify the model's predicted class.

\subsubsection{Counterfactual subset}

The definition of a counterfactual explanation includes the requirement of having an irreducible set of valid modifications (as described in Equation~\ref{eq:irred}). Although the interpretation of valid modifications is debatable~\cite{de2021framework}, being able to identify subsets of changes that form a counterfactual explanation may be useful for a better understanding of the prediction model and for the assessment of the counterfactual explanation.

\begin{figure}[h]
\centering
\includegraphics[width=12cm]{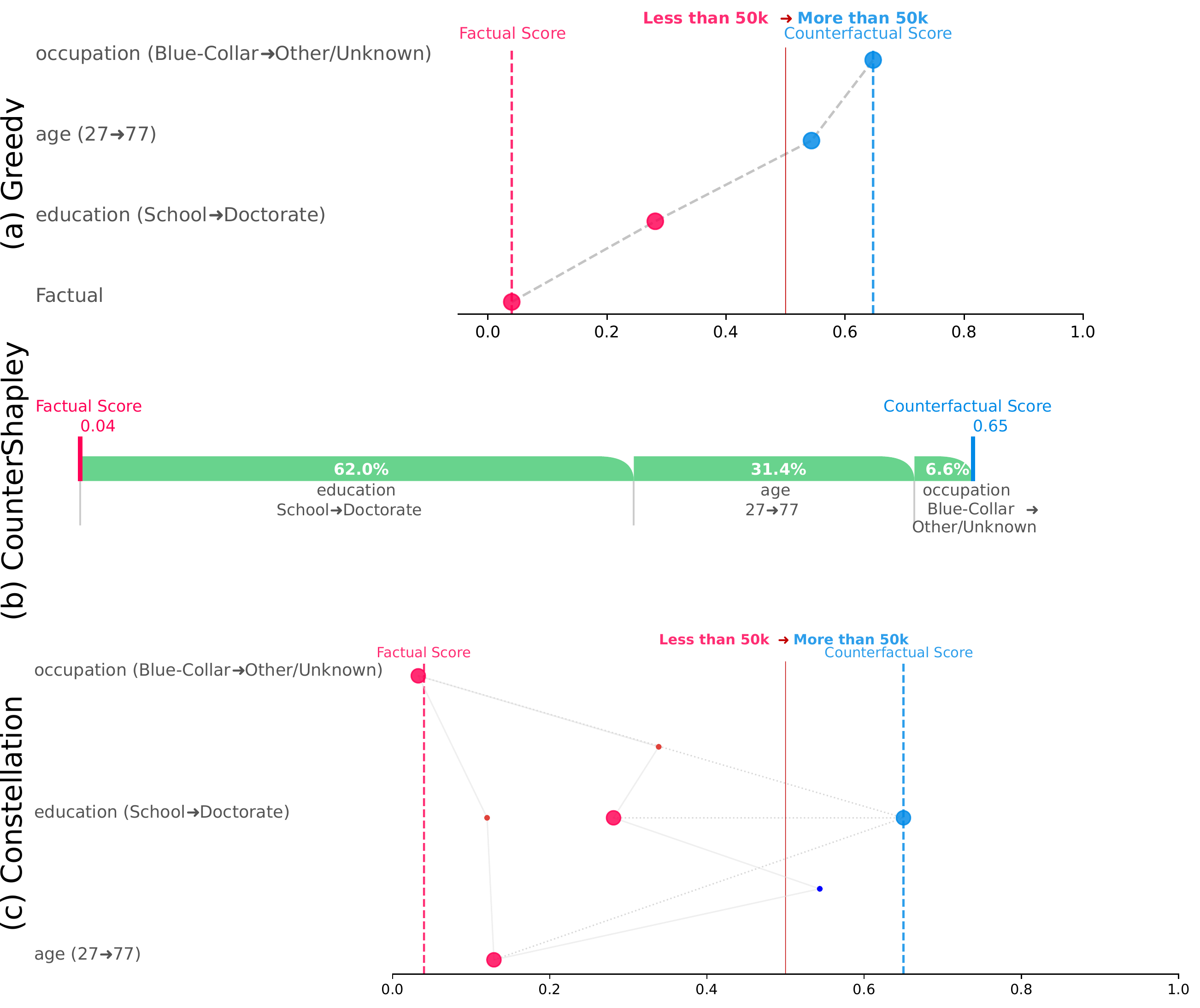}
\caption{Example of a counterfactual with a subset that also flips the prediction class. We use the Adult dataset and the DiCE counterfactual generator for this case.}
\label{fig:examplesubset}
\end{figure}

The usual simple counterfactual results are not capable of showing this since they only show the feature changes needed for a class modification. Therefore, the analysis presented in this work may be a valuable tool for evaluating the counterfactual result. Figure~\ref{fig:examplesubset} shows an example of how a counterfactual with class changing subsets would look in Greedy, CounterShapley, and Constellation charts.

In our example figure, the Greedy and CounterShapley charts indicate that the counterfactual explanation has more features than needed to change the classification change. The Constellation Chart may be the most appropriate analysis since it clearly shows which subsets of change lead to a counterfactual class. For this example, we can see the change in \textit{occupation} feature is not needed to modify the prediction class.

\newpage

\section{Conclusion}

Machine learning predictions can be effectively explained with counterfactual explanations. Its simple and decision-driven nature makes it easier for even laypersons to get a sufficient understanding of why automated decisions are made. However, up to now, the representation of this type of explanation was mainly based on textual communication which, although has its own benefits, lacks the benefits of graphical elements. Moreover, the isonomic view of the counterfactual modified features may not give a complete picture of why a model has certain decisions since it does not explain the relationship between features and prediction scoring. This paper proposes two CFI methods as a solution to those challenges,  Greedy and CounterShapley, as being approaches that can embed relative importance to counterfactual features (and their corresponding value modifications) and three chart types that highlight different aspects of the counterfactual explanation. Hence, the Greedy CFI method illustrates how a specific path, following the steepest increases of scores, links to the counterfactual changes, and CounterShapley fills the knowledge gap of how the prediction scores are affected by each changed feature in overall. Those CFI methods can provide better insights into how the model works, consequently augmenting the informative value of the explanations. Regarding the plotting strategies, CounterShapley charts use the feature change importance values to create simple yet descriptive graphs. For the cases in which computational burden is an issue, we propose the Greedy Counterfactual chart to quickly provide a visual representation that describes the greediest strategy to achieve the counterfactual prediction score. The third chart type, the Constellation chart, takes advantage of the sparse nature of counterfactual explanations to show the possible feature associations and their impact on the prediction scoring. Despite the usefulness of these methods in augmenting the informative value of counterfactual explanations, they are not adequate for situations where the characteristics of counterfactuals are not desirable, for example, if an importance score is required for every model's feature. Finally, all methods and charts here described are implemented in an open-source package that can be easily integrated into any counterfactual generator, bringing immediate impact considering the multiple counterfactual generation algorithms available. As for future perspective, we envision the application of counterfactual-based scores and their visual representations in real-world cases, assessing the user reception in terms of understandability and preference over other explainability approaches.

\bibliography{bib}

\begin{thebibliography}{53}
\expandafter\ifx\csname natexlab\endcsname\relax\def\natexlab#1{#1}\fi
\providecommand{\url}[1]{\texttt{#1}}
\providecommand{\href}[2]{#2}
\providecommand{\path}[1]{#1}
\providecommand{\DOIprefix}{doi:}
\providecommand{\ArXivprefix}{arXiv:}
\providecommand{\URLprefix}{URL: }
\providecommand{\Pubmedprefix}{pmid:}
\providecommand{\doi}[1]{\href{http://dx.doi.org/#1}{\path{#1}}}
\providecommand{\Pubmed}[1]{\href{pmid:#1}{\path{#1}}}
\providecommand{\bibinfo}[2]{#2}
\ifx\xfnm\relax \def\xfnm[#1]{\unskip,\space#1}\fi
%Type = Article
\bibitem[{Brown et~al.(2020)Brown, Mann, Ryder, Subbiah, Kaplan, Dhariwal,
  Neelakantan, Shyam, Sastry, Askell et~al.}]{brown2020language}
\bibinfo{author}{T.~Brown}, \bibinfo{author}{B.~Mann},
  \bibinfo{author}{N.~Ryder}, \bibinfo{author}{M.~Subbiah},
  \bibinfo{author}{J.~D. Kaplan}, \bibinfo{author}{P.~Dhariwal},
  \bibinfo{author}{A.~Neelakantan}, \bibinfo{author}{P.~Shyam},
  \bibinfo{author}{G.~Sastry}, \bibinfo{author}{A.~Askell}, et~al.,
\newblock \bibinfo{title}{Language models are few-shot learners},
\newblock \bibinfo{journal}{Advances in neural information processing systems}
  \bibinfo{volume}{33} (\bibinfo{year}{2020}) \bibinfo{pages}{1877--1901}.
%Type = Inproceedings
\bibitem[{Ronanki et~al.(2016)Ronanki, Henter, Wu, and
  King}]{ronanki2016template}
\bibinfo{author}{S.~Ronanki}, \bibinfo{author}{G.~E. Henter},
  \bibinfo{author}{Z.~Wu}, \bibinfo{author}{S.~King},
\newblock \bibinfo{title}{A template-based approach for speech synthesis
  intonation generation using lstms.},
\newblock in: \bibinfo{booktitle}{INTERSPEECH}, \bibinfo{year}{2016}, pp.
  \bibinfo{pages}{2463--2467}.
%Type = Inproceedings
\bibitem[{Neekhara et~al.(2021)Neekhara, Hussain, Dubnov, Koushanfar, and
  McAuley}]{neekhara2021expressive}
\bibinfo{author}{P.~Neekhara}, \bibinfo{author}{S.~Hussain},
  \bibinfo{author}{S.~Dubnov}, \bibinfo{author}{F.~Koushanfar},
  \bibinfo{author}{J.~McAuley},
\newblock \bibinfo{title}{Expressive neural voice cloning},
\newblock in: \bibinfo{booktitle}{Asian Conference on Machine Learning},
  \bibinfo{organization}{PMLR}, \bibinfo{year}{2021}, pp.
  \bibinfo{pages}{252--267}.
%Type = Article
\bibitem[{Yang and Chung(2019)}]{yang2019self}
\bibinfo{author}{S.~H. Yang}, \bibinfo{author}{M.~Chung},
\newblock \bibinfo{title}{Self-imitating feedback generation using gan for
  computer-assisted pronunciation training},
\newblock \bibinfo{journal}{arXiv preprint arXiv:1904.09407}
  (\bibinfo{year}{2019}).
%Type = Article
\bibitem[{Arrieta et~al.(2020)Arrieta, D{\'\i}az-Rodr{\'\i}guez, Del~Ser,
  Bennetot, Tabik, Barbado, Garc{\'\i}a, Gil-L{\'o}pez, Molina, Benjamins
  et~al.}]{arrieta2020explainable}
\bibinfo{author}{A.~B. Arrieta}, \bibinfo{author}{N.~D{\'\i}az-Rodr{\'\i}guez},
  \bibinfo{author}{J.~Del~Ser}, \bibinfo{author}{A.~Bennetot},
  \bibinfo{author}{S.~Tabik}, \bibinfo{author}{A.~Barbado},
  \bibinfo{author}{S.~Garc{\'\i}a}, \bibinfo{author}{S.~Gil-L{\'o}pez},
  \bibinfo{author}{D.~Molina}, \bibinfo{author}{R.~Benjamins}, et~al.,
\newblock \bibinfo{title}{Explainable artificial intelligence (xai): Concepts,
  taxonomies, opportunities and challenges toward responsible ai},
\newblock \bibinfo{journal}{Information fusion} \bibinfo{volume}{58}
  (\bibinfo{year}{2020}) \bibinfo{pages}{82--115}.
%Type = Article
\bibitem[{Meske et~al.(2022)Meske, Bunde, Schneider, and
  Gersch}]{meske2022explainable}
\bibinfo{author}{C.~Meske}, \bibinfo{author}{E.~Bunde},
  \bibinfo{author}{J.~Schneider}, \bibinfo{author}{M.~Gersch},
\newblock \bibinfo{title}{Explainable artificial intelligence: objectives,
  stakeholders, and future research opportunities},
\newblock \bibinfo{journal}{Information Systems Management}
  \bibinfo{volume}{39} (\bibinfo{year}{2022}) \bibinfo{pages}{53--63}.
%Type = Article
\bibitem[{Wachter et~al.(2017)Wachter, Mittelstadt, and
  Russell}]{wachter2017counterfactual}
\bibinfo{author}{S.~Wachter}, \bibinfo{author}{B.~Mittelstadt},
  \bibinfo{author}{C.~Russell},
\newblock \bibinfo{title}{Counterfactual explanations without opening the black
  box: Automated decisions and the gdpr},
\newblock \bibinfo{journal}{Harv. JL \& Tech.} \bibinfo{volume}{31}
  (\bibinfo{year}{2017}) \bibinfo{pages}{841}.
%Type = Article
\bibitem[{Goethals et~al.(2022)Goethals, Martens, and
  Calders}]{goethals2022precof}
\bibinfo{author}{S.~Goethals}, \bibinfo{author}{D.~Martens},
  \bibinfo{author}{T.~Calders},
\newblock \bibinfo{title}{Precof: Counterfactual explanations for fairness}
  (\bibinfo{year}{2022}).
%Type = Book
\bibitem[{Martens(2022)}]{martensdsethics2022}
\bibinfo{author}{D.~Martens}, \bibinfo{title}{Data Science Ethics: Concepts,
  Techniques, and Cautionary Tales}, \bibinfo{publisher}{Oxford University
  Press}, \bibinfo{year}{2022}.
  \DOIprefix\doi{10.1093/oso/9780192847263.001.0001}.
%Type = Inproceedings
\bibitem[{Speith(2022)}]{speith2022review}
\bibinfo{author}{T.~Speith},
\newblock \bibinfo{title}{A review of taxonomies of explainable artificial
  intelligence (xai) methods},
\newblock in: \bibinfo{booktitle}{2022 ACM Conference on Fairness,
  Accountability, and Transparency}, \bibinfo{year}{2022}, pp.
  \bibinfo{pages}{2239--2250}.
%Type = Article
\bibitem[{Breiman(2001)}]{breiman2001random}
\bibinfo{author}{L.~Breiman},
\newblock \bibinfo{title}{Random forests},
\newblock \bibinfo{journal}{Machine learning} \bibinfo{volume}{45}
  (\bibinfo{year}{2001}) \bibinfo{pages}{5--32}.
%Type = Misc
\bibitem[{Craven and Shavlik(1996)}]{craven1996extracting}
\bibinfo{author}{M.~Craven}, \bibinfo{author}{J.~Shavlik},
  \bibinfo{title}{Extracting tree-structured representations of trained
  networks, advances, neural information processing systems 8},
  \bibinfo{year}{1996}.
%Type = Article
\bibitem[{{Martens} et~al.(2009){Martens}, {Baesens}, and {Van
  Gestel}}]{martensalba2009}
\bibinfo{author}{D.~{Martens}}, \bibinfo{author}{B.~{Baesens}},
  \bibinfo{author}{T.~{Van Gestel}},
\newblock \bibinfo{title}{Decompositional rule extraction from support vector
  machines by active learning},
\newblock \bibinfo{journal}{IEEE Transactions on Knowledge and Data
  Engineering} \bibinfo{volume}{21} (\bibinfo{year}{2009})
  \bibinfo{pages}{178--191}.
%Type = Book
\bibitem[{Hastie et~al.(2001)Hastie, Tibshirani, and
  Friedman}]{hastie01statisticallearning}
\bibinfo{author}{T.~Hastie}, \bibinfo{author}{R.~Tibshirani},
  \bibinfo{author}{J.~Friedman}, \bibinfo{title}{The Elements of Statistical
  Learning}, Springer Series in Statistics, \bibinfo{publisher}{Springer New
  York Inc.}, \bibinfo{address}{New York, NY, USA}, \bibinfo{year}{2001}.
%Type = Inproceedings
\bibitem[{Ribeiro et~al.(2016)Ribeiro, Singh, and Guestrin}]{lime}
\bibinfo{author}{M.~T. Ribeiro}, \bibinfo{author}{S.~Singh},
  \bibinfo{author}{C.~Guestrin},
\newblock \bibinfo{title}{"why should {I} trust you?": Explaining the
  predictions of any classifier},
\newblock in: \bibinfo{booktitle}{Proceedings of the 22nd {ACM} {SIGKDD}
  International Conference on Knowledge Discovery and Data Mining, San
  Francisco, CA, USA, August 13-17, 2016}, \bibinfo{year}{2016}, pp.
  \bibinfo{pages}{1135--1144}.
%Type = Article
\bibitem[{Lundberg and Lee(2017)}]{lundberg2017unified}
\bibinfo{author}{S.~M. Lundberg}, \bibinfo{author}{S.-I. Lee},
\newblock \bibinfo{title}{A unified approach to interpreting model
  predictions},
\newblock \bibinfo{journal}{Advances in neural information processing systems}
  \bibinfo{volume}{30} (\bibinfo{year}{2017}).
%Type = Article
\bibitem[{Verma et~al.(2020)Verma, Dickerson, and
  Hines}]{verma2020counterfactual}
\bibinfo{author}{S.~Verma}, \bibinfo{author}{J.~Dickerson},
  \bibinfo{author}{K.~Hines},
\newblock \bibinfo{title}{Counterfactual explanations for machine learning: A
  review},
\newblock \bibinfo{journal}{arXiv preprint arXiv:2010.10596}
  (\bibinfo{year}{2020}).
%Type = Inproceedings
\bibitem[{Buitinck et~al.(2013)Buitinck, Louppe, Blondel, Pedregosa, Mueller,
  Grisel, Niculae, Prettenhofer, Gramfort, Grobler, Layton, VanderPlas, Joly,
  Holt, and Varoquaux}]{sklearn_api}
\bibinfo{author}{L.~Buitinck}, \bibinfo{author}{G.~Louppe},
  \bibinfo{author}{M.~Blondel}, \bibinfo{author}{F.~Pedregosa},
  \bibinfo{author}{A.~Mueller}, \bibinfo{author}{O.~Grisel},
  \bibinfo{author}{V.~Niculae}, \bibinfo{author}{P.~Prettenhofer},
  \bibinfo{author}{A.~Gramfort}, \bibinfo{author}{J.~Grobler},
  \bibinfo{author}{R.~Layton}, \bibinfo{author}{J.~VanderPlas},
  \bibinfo{author}{A.~Joly}, \bibinfo{author}{B.~Holt},
  \bibinfo{author}{G.~Varoquaux},
\newblock \bibinfo{title}{{API} design for machine learning software:
  experiences from the scikit-learn project},
\newblock in: \bibinfo{booktitle}{ECML PKDD Workshop: Languages for Data Mining
  and Machine Learning}, \bibinfo{year}{2013}, pp. \bibinfo{pages}{108--122}.
%Type = Article
\bibitem[{Shapley(1953)}]{shapley1953quota}
\bibinfo{author}{L.~Shapley},
\newblock \bibinfo{title}{Quota solutions op n-person games1},
\newblock \bibinfo{journal}{Edited by Emil Artin and Marston Morse}
  (\bibinfo{year}{1953}) \bibinfo{pages}{343}.
%Type = Article
\bibitem[{Stepin et~al.(2021)Stepin, Alonso, Catala, and
  Pereira-Fari{\~n}a}]{stepin2021survey}
\bibinfo{author}{I.~Stepin}, \bibinfo{author}{J.~M. Alonso},
  \bibinfo{author}{A.~Catala}, \bibinfo{author}{M.~Pereira-Fari{\~n}a},
\newblock \bibinfo{title}{A survey of contrastive and counterfactual
  explanation generation methods for explainable artificial intelligence},
\newblock \bibinfo{journal}{IEEE Access} \bibinfo{volume}{9}
  (\bibinfo{year}{2021}) \bibinfo{pages}{11974--12001}.
%Type = Article
\bibitem[{Martens and Provost(2011)}]{martens2011explaining}
\bibinfo{author}{D.~Martens}, \bibinfo{author}{F.~Provost},
\newblock \bibinfo{title}{Explaining documents’ classifications},
\newblock \bibinfo{journal}{Center for Digital Economy Research}
  (\bibinfo{year}{2011}).
%Type = Article
\bibitem[{de~Oliveira and Martens(2021)}]{de2021framework}
\bibinfo{author}{R.~M.~B. de~Oliveira}, \bibinfo{author}{D.~Martens},
\newblock \bibinfo{title}{A framework and benchmarking study for counterfactual
  generating methods on tabular data},
\newblock \bibinfo{journal}{Applied Sciences} \bibinfo{volume}{11}
  (\bibinfo{year}{2021}) \bibinfo{pages}{7274}.
%Type = Article
\bibitem[{Fern{\'a}ndez-Lor{\'\i}a et~al.(2020)Fern{\'a}ndez-Lor{\'\i}a,
  Provost, and Han}]{fernandez2020explaining}
\bibinfo{author}{C.~Fern{\'a}ndez-Lor{\'\i}a}, \bibinfo{author}{F.~Provost},
  \bibinfo{author}{X.~Han},
\newblock \bibinfo{title}{Explaining data-driven decisions made by ai systems:
  The counterfactual approach},
\newblock \bibinfo{journal}{arXiv preprint arXiv:2001.07417}
  (\bibinfo{year}{2020}).
%Type = Article
\bibitem[{Byrne(2007)}]{byrne2007precis}
\bibinfo{author}{R.~M. Byrne},
\newblock \bibinfo{title}{Precis of the rational imagination: How people create
  alternatives to reality},
\newblock \bibinfo{journal}{Behavioral and Brain Sciences} \bibinfo{volume}{30}
  (\bibinfo{year}{2007}) \bibinfo{pages}{439--453}.
%Type = Article
\bibitem[{Lipton(1990)}]{lipton1990contrastive}
\bibinfo{author}{P.~Lipton},
\newblock \bibinfo{title}{Contrastive explanation},
\newblock \bibinfo{journal}{Royal Institute of Philosophy Supplements}
  \bibinfo{volume}{27} (\bibinfo{year}{1990}) \bibinfo{pages}{247--266}.
%Type = Article
\bibitem[{Vermeire et~al.(2022)Vermeire, Brughmans, Goethals, de~Oliveira, and
  Martens}]{vermeire2022explainable}
\bibinfo{author}{T.~Vermeire}, \bibinfo{author}{D.~Brughmans},
  \bibinfo{author}{S.~Goethals}, \bibinfo{author}{R.~M.~B. de~Oliveira},
  \bibinfo{author}{D.~Martens},
\newblock \bibinfo{title}{Explainable image classification with evidence
  counterfactual},
\newblock \bibinfo{journal}{Pattern Analysis and Applications}
  (\bibinfo{year}{2022}) \bibinfo{pages}{1--21}.
%Type = Article
\bibitem[{Ramon et~al.(2021)Ramon, Vermeire, Toubia, Martens, and
  Evgeniou}]{ramon2021understanding}
\bibinfo{author}{Y.~Ramon}, \bibinfo{author}{T.~Vermeire},
  \bibinfo{author}{O.~Toubia}, \bibinfo{author}{D.~Martens},
  \bibinfo{author}{T.~Evgeniou},
\newblock \bibinfo{title}{Understanding consumer preferences for explanations
  generated by xai algorithms},
\newblock \bibinfo{journal}{arXiv preprint arXiv:2107.02624}
  (\bibinfo{year}{2021}).
%Type = Article
\bibitem[{Martens and Provost(2014)}]{martens2014explaining}
\bibinfo{author}{D.~Martens}, \bibinfo{author}{F.~Provost},
\newblock \bibinfo{title}{Explaining data-driven document classifications},
\newblock \bibinfo{journal}{MIS quarterly} \bibinfo{volume}{38}
  (\bibinfo{year}{2014}) \bibinfo{pages}{73--100}.
%Type = Inproceedings
\bibitem[{Hohman et~al.(2019)Hohman, Head, Caruana, DeLine, and
  Drucker}]{hohman2019gamut}
\bibinfo{author}{F.~Hohman}, \bibinfo{author}{A.~Head},
  \bibinfo{author}{R.~Caruana}, \bibinfo{author}{R.~DeLine},
  \bibinfo{author}{S.~M. Drucker},
\newblock \bibinfo{title}{Gamut: A design probe to understand how data
  scientists understand machine learning models},
\newblock in: \bibinfo{booktitle}{Proceedings of the 2019 CHI conference on
  human factors in computing systems}, \bibinfo{year}{2019}, pp.
  \bibinfo{pages}{1--13}.
%Type = Article
\bibitem[{Wexler et~al.(2019)Wexler, Pushkarna, Bolukbasi, Wattenberg,
  Vi{\'e}gas, and Wilson}]{wexler2019if}
\bibinfo{author}{J.~Wexler}, \bibinfo{author}{M.~Pushkarna},
  \bibinfo{author}{T.~Bolukbasi}, \bibinfo{author}{M.~Wattenberg},
  \bibinfo{author}{F.~Vi{\'e}gas}, \bibinfo{author}{J.~Wilson},
\newblock \bibinfo{title}{The what-if tool: Interactive probing of machine
  learning models},
\newblock \bibinfo{journal}{IEEE transactions on visualization and computer
  graphics} \bibinfo{volume}{26} (\bibinfo{year}{2019})
  \bibinfo{pages}{56--65}.
%Type = Article
\bibitem[{Cheng et~al.(2020)Cheng, Ming, and Qu}]{cheng2020dece}
\bibinfo{author}{F.~Cheng}, \bibinfo{author}{Y.~Ming}, \bibinfo{author}{H.~Qu},
\newblock \bibinfo{title}{Dece: Decision explorer with counterfactual
  explanations for machine learning models},
\newblock \bibinfo{journal}{IEEE Transactions on Visualization and Computer
  Graphics} \bibinfo{volume}{27} (\bibinfo{year}{2020})
  \bibinfo{pages}{1438--1447}.
%Type = Inproceedings
\bibitem[{Gomez et~al.(2020)Gomez, Holter, Yuan, and Bertini}]{gomez2020vice}
\bibinfo{author}{O.~Gomez}, \bibinfo{author}{S.~Holter},
  \bibinfo{author}{J.~Yuan}, \bibinfo{author}{E.~Bertini},
\newblock \bibinfo{title}{Vice: visual counterfactual explanations for machine
  learning models},
\newblock in: \bibinfo{booktitle}{Proceedings of the 25th International
  Conference on Intelligent User Interfaces}, \bibinfo{year}{2020}, pp.
  \bibinfo{pages}{531--535}.
%Type = Article
\bibitem[{Angelov et~al.(2021)Angelov, Soares, Jiang, Arnold, and
  Atkinson}]{angelov2021explainable}
\bibinfo{author}{P.~P. Angelov}, \bibinfo{author}{E.~A. Soares},
  \bibinfo{author}{R.~Jiang}, \bibinfo{author}{N.~I. Arnold},
  \bibinfo{author}{P.~M. Atkinson},
\newblock \bibinfo{title}{Explainable artificial intelligence: an analytical
  review},
\newblock \bibinfo{journal}{Wiley Interdisciplinary Reviews: Data Mining and
  Knowledge Discovery} \bibinfo{volume}{11} (\bibinfo{year}{2021})
  \bibinfo{pages}{e1424}.
%Type = Article
\bibitem[{Winter(2002)}]{winter2002shapley}
\bibinfo{author}{E.~Winter},
\newblock \bibinfo{title}{The shapley value},
\newblock \bibinfo{journal}{Handbook of game theory with economic applications}
  \bibinfo{volume}{3} (\bibinfo{year}{2002}) \bibinfo{pages}{2025--2054}.
%Type = Article
\bibitem[{Strumbelj and Kononenko(2010)}]{strumbelj2010efficient}
\bibinfo{author}{E.~Strumbelj}, \bibinfo{author}{I.~Kononenko},
\newblock \bibinfo{title}{An efficient explanation of individual
  classifications using game theory},
\newblock \bibinfo{journal}{The Journal of Machine Learning Research}
  \bibinfo{volume}{11} (\bibinfo{year}{2010}) \bibinfo{pages}{1--18}.
%Type = Article
\bibitem[{Zafar and Khan(2021)}]{zafar2021deterministic}
\bibinfo{author}{M.~R. Zafar}, \bibinfo{author}{N.~Khan},
\newblock \bibinfo{title}{Deterministic local interpretable model-agnostic
  explanations for stable explainability},
\newblock \bibinfo{journal}{Machine Learning and Knowledge Extraction}
  \bibinfo{volume}{3} (\bibinfo{year}{2021}) \bibinfo{pages}{525--541}.
%Type = Article
\bibitem[{Visani et~al.(2022)Visani, Bagli, Chesani, Poluzzi, and
  Capuzzo}]{visani2022statistical}
\bibinfo{author}{G.~Visani}, \bibinfo{author}{E.~Bagli},
  \bibinfo{author}{F.~Chesani}, \bibinfo{author}{A.~Poluzzi},
  \bibinfo{author}{D.~Capuzzo},
\newblock \bibinfo{title}{Statistical stability indices for lime: Obtaining
  reliable explanations for machine learning models},
\newblock \bibinfo{journal}{Journal of the Operational Research Society}
  \bibinfo{volume}{73} (\bibinfo{year}{2022}) \bibinfo{pages}{91--101}.
%Type = Article
\bibitem[{Zafar and Khan(2019)}]{zafar2019dlime}
\bibinfo{author}{M.~R. Zafar}, \bibinfo{author}{N.~M. Khan},
\newblock \bibinfo{title}{Dlime: A deterministic local interpretable
  model-agnostic explanations approach for computer-aided diagnosis systems},
\newblock \bibinfo{journal}{arXiv preprint arXiv:1906.10263}
  (\bibinfo{year}{2019}).
%Type = Article
\bibitem[{{\v{S}}trumbelj and Kononenko(2014)}]{vstrumbelj2014explaining}
\bibinfo{author}{E.~{\v{S}}trumbelj}, \bibinfo{author}{I.~Kononenko},
\newblock \bibinfo{title}{Explaining prediction models and individual
  predictions with feature contributions},
\newblock \bibinfo{journal}{Knowledge and information systems}
  \bibinfo{volume}{41} (\bibinfo{year}{2014}) \bibinfo{pages}{647--665}.
%Type = Article
\bibitem[{Lundberg et~al.(2020)Lundberg, Erion, Chen, DeGrave, Prutkin, Nair,
  Katz, Himmelfarb, Bansal, and Lee}]{10.1038/s42256-019-0138-9}
\bibinfo{author}{S.~M. Lundberg}, \bibinfo{author}{G.~Erion},
  \bibinfo{author}{H.~Chen}, \bibinfo{author}{A.~DeGrave},
  \bibinfo{author}{J.~M. Prutkin}, \bibinfo{author}{B.~Nair},
  \bibinfo{author}{R.~Katz}, \bibinfo{author}{J.~Himmelfarb},
  \bibinfo{author}{N.~Bansal}, \bibinfo{author}{S.-I. Lee},
\newblock \bibinfo{title}{{From local explanations to global understanding with
  explainable AI for trees}},
\newblock \bibinfo{journal}{Nature Machine Intelligence} \bibinfo{volume}{2}
  (\bibinfo{year}{2020}) \bibinfo{pages}{56--67}.
%Type = Article
\bibitem[{Ghalebikesabi et~al.(2021)Ghalebikesabi, Ter-Minassian, DiazOrdaz,
  and Holmes}]{ghalebikesabi2021locality}
\bibinfo{author}{S.~Ghalebikesabi}, \bibinfo{author}{L.~Ter-Minassian},
  \bibinfo{author}{K.~DiazOrdaz}, \bibinfo{author}{C.~C. Holmes},
\newblock \bibinfo{title}{On locality of local explanation models},
\newblock \bibinfo{journal}{Advances in Neural Information Processing Systems}
  \bibinfo{volume}{34} (\bibinfo{year}{2021}) \bibinfo{pages}{18395--18407}.
%Type = Inproceedings
\bibitem[{Moore et~al.(2019)Moore, Hammerla, and Watkins}]{moore2019explaining}
\bibinfo{author}{J.~Moore}, \bibinfo{author}{N.~Hammerla},
  \bibinfo{author}{C.~Watkins},
\newblock \bibinfo{title}{Explaining deep learning models with constrained
  adversarial examples},
\newblock in: \bibinfo{booktitle}{Pacific Rim international conference on
  artificial intelligence}, \bibinfo{organization}{Springer},
  \bibinfo{year}{2019}, pp. \bibinfo{pages}{43--56}.
%Type = Inproceedings
\bibitem[{Mothilal et~al.(2020)Mothilal, Sharma, and
  Tan}]{mothilal2020explaining}
\bibinfo{author}{R.~K. Mothilal}, \bibinfo{author}{A.~Sharma},
  \bibinfo{author}{C.~Tan},
\newblock \bibinfo{title}{Explaining machine learning classifiers through
  diverse counterfactual explanations},
\newblock in: \bibinfo{booktitle}{Proceedings of the 2020 conference on
  fairness, accountability, and transparency}, \bibinfo{year}{2020}, pp.
  \bibinfo{pages}{607--617}.
%Type = Inproceedings
\bibitem[{Looveren and Klaise(2021)}]{looveren2021interpretable}
\bibinfo{author}{A.~V. Looveren}, \bibinfo{author}{J.~Klaise},
\newblock \bibinfo{title}{Interpretable counterfactual explanations guided by
  prototypes},
\newblock in: \bibinfo{booktitle}{Joint European Conference on Machine Learning
  and Knowledge Discovery in Databases}, \bibinfo{organization}{Springer},
  \bibinfo{year}{2021}, pp. \bibinfo{pages}{650--665}.
%Type = Article
\bibitem[{Ainsworth and Th~Loizou(2003)}]{ainsworth2003effects}
\bibinfo{author}{S.~Ainsworth}, \bibinfo{author}{A.~Th~Loizou},
\newblock \bibinfo{title}{The effects of self-explaining when learning with
  text or diagrams},
\newblock \bibinfo{journal}{Cognitive science} \bibinfo{volume}{27}
  (\bibinfo{year}{2003}) \bibinfo{pages}{669--681}.
%Type = Article
\bibitem[{Brughmans et~al.(2021)Brughmans, Leyman, and
  Martens}]{brughmans2021nice}
\bibinfo{author}{D.~Brughmans}, \bibinfo{author}{P.~Leyman},
  \bibinfo{author}{D.~Martens},
\newblock \bibinfo{title}{Nice: an algorithm for nearest instance
  counterfactual explanations},
\newblock \bibinfo{journal}{arXiv preprint arXiv:2104.07411}
  (\bibinfo{year}{2021}).
%Type = Article
\bibitem[{Vilone and Longo(2021)}]{vilone2021notions}
\bibinfo{author}{G.~Vilone}, \bibinfo{author}{L.~Longo},
\newblock \bibinfo{title}{Notions of explainability and evaluation approaches
  for explainable artificial intelligence},
\newblock \bibinfo{journal}{Information Fusion} \bibinfo{volume}{76}
  (\bibinfo{year}{2021}) \bibinfo{pages}{89--106}.
%Type = Article
\bibitem[{Hunter(2007)}]{Hunter:2007}
\bibinfo{author}{J.~D. Hunter},
\newblock \bibinfo{title}{Matplotlib: A 2d graphics environment},
\newblock \bibinfo{journal}{Computing in Science \& Engineering}
  \bibinfo{volume}{9} (\bibinfo{year}{2007}) \bibinfo{pages}{90--95}.
%Type = Article
\bibitem[{Pedregosa et~al.(2011)Pedregosa, Varoquaux, Gramfort, Michel,
  Thirion, Grisel, Blondel, Prettenhofer, Weiss, Dubourg, Vanderplas, Passos,
  Cournapeau, Brucher, Perrot, and Duchesnay}]{scikit-learn}
\bibinfo{author}{F.~Pedregosa}, \bibinfo{author}{G.~Varoquaux},
  \bibinfo{author}{A.~Gramfort}, \bibinfo{author}{V.~Michel},
  \bibinfo{author}{B.~Thirion}, \bibinfo{author}{O.~Grisel},
  \bibinfo{author}{M.~Blondel}, \bibinfo{author}{P.~Prettenhofer},
  \bibinfo{author}{R.~Weiss}, \bibinfo{author}{V.~Dubourg},
  \bibinfo{author}{J.~Vanderplas}, \bibinfo{author}{A.~Passos},
  \bibinfo{author}{D.~Cournapeau}, \bibinfo{author}{M.~Brucher},
  \bibinfo{author}{M.~Perrot}, \bibinfo{author}{E.~Duchesnay},
\newblock \bibinfo{title}{Scikit-learn: Machine learning in {P}ython},
\newblock \bibinfo{journal}{Journal of Machine Learning Research}
  \bibinfo{volume}{12} (\bibinfo{year}{2011}) \bibinfo{pages}{2825--2830}.
%Type = Misc
\bibitem[{pandas~development team(2020)}]{reback2020pandas}
\bibinfo{author}{T.~pandas~development team},
  \bibinfo{title}{pandas-dev/pandas: Pandas}, \bibinfo{year}{2020}. \URLprefix
  \url{https://doi.org/10.5281/zenodo.3509134}.
  \DOIprefix\doi{10.5281/zenodo.3509134}.
%Type = Article
\bibitem[{Waskom(2021)}]{Waskom2021}
\bibinfo{author}{M.~L. Waskom},
\newblock \bibinfo{title}{seaborn: statistical data visualization},
\newblock \bibinfo{journal}{Journal of Open Source Software}
  \bibinfo{volume}{6} (\bibinfo{year}{2021}) \bibinfo{pages}{3021}.
%Type = Article
\bibitem[{Robnik-{\v{S}}ikonja and Kononenko(2008)}]{robnik2008explaining}
\bibinfo{author}{M.~Robnik-{\v{S}}ikonja}, \bibinfo{author}{I.~Kononenko},
\newblock \bibinfo{title}{Explaining classifications for individual instances},
\newblock \bibinfo{journal}{IEEE Transactions on Knowledge and Data
  Engineering} \bibinfo{volume}{20} (\bibinfo{year}{2008})
  \bibinfo{pages}{589--600}.
%Type = Article
\bibitem[{Liang et~al.(2016)Liang, Lu, Tsai, and Shih}]{liang2016financial}
\bibinfo{author}{D.~Liang}, \bibinfo{author}{C.-C. Lu}, \bibinfo{author}{C.-F.
  Tsai}, \bibinfo{author}{G.-A. Shih},
\newblock \bibinfo{title}{Financial ratios and corporate governance indicators
  in bankruptcy prediction: A comprehensive study},
\newblock \bibinfo{journal}{European journal of operational research}
  \bibinfo{volume}{252} (\bibinfo{year}{2016}) \bibinfo{pages}{561--572}.

\end{thebibliography}

\begin{appendices}

\LinesNotNumbered{
\begin{algorithm}
\label{niceexample}
\caption{NICE's Python package adaptation for integrating the article's methods}
\begin{python_nobox}
from nice import NICE
from counterplots import CreatePlot

class CounterPlotNice(NICE):

    def __init__(self, predict_fn, *args, **kwargs):
        super().__init__(predict_fn, *args, **kwargs)
        self.predict_fn = predict_fn

    def explain(self, X, target_class='other'):
        explanations = super().explain(X, target_class)

        out_exp = []

        for i in range(len(X)):
            out_exp.append(
                CreatePlot(factual=X[i], 
                cf=explanations[i], 
                model_pred=self.predict_fn)
            )

        return out_exp
\end{python_nobox}
\end{algorithm}
}

\LinesNotNumbered{
\begin{algorithm}[h]
\label{diceexample}
\caption{Function which takes DiCE counterfactual results and generates this article's analysis and plots}
\begin{python_nobox}
import json
import pandas as pd
from dice_ml import Dice
from counterplots import CreatePlot

# exp and dice_exp is generated using DiCE package, following official documentation

def dice_counterplots(dice_exp, exp):
    
    cf_data = json.loads(dice_exp.to_json())
    factual = cf_data['test_data'][0][0][:-1]
    feature_names = exp.data_interface.feature_names
    df_structure = exp.data_interface.data_df[:0].loc[:, feature_names]
    data_types = df_structure.dtypes.apply(lambda x: x.name).to_dict()
    
    def adjust_types(x):
        for i in range(x.shape[1]):
            if 'int' in list(data_types.values())[i]:
                x[:, i] = int(float(x[:, i]))
        return x
    
    def model_pred(x):
        scores = exp.predict_fn(df_structure.append(pd.DataFrame(x, columns=feature_names))).numpy()
        return np.concatenate((1 - scores, scores), axis=1)
    
    out_exp = []
    
    for raw_cf in cf_data['cfs_list'][0]:
        cf = adjust_types(np.array([raw_cf[:-1]]))[0]
    
        out_exp.append(
            CreatePlot(
                factual=np.array(factual),
                cf=np.array(cf),
                model_pred=model_pred,
                feature_names=feature_names)
            )
    
    return out_exp
\end{python_nobox}
\end{algorithm}
}

\LinesNotNumbered{
\begin{algorithm}
\label{alibicexample}
\caption{Function which takes DiCE counterfactual results and generates this article's analysis and plots}
\begin{python_nobox}
from alibi.explainers import CounterfactualProto
from counterplots import CreatePlot

# factual and model are obtained from the dataset treatment/training
# and exp is created with the instance and model data using ALIBIC package

def CounterPlotAlibic(factual, exp, model):
    cf = exp.cf['X'][0]
    
    out_exp = CreatePlot(
        factual=factual,
        cf=cf,
        model_pred=model
    )
    
    return out_exp
\end{python_nobox}
\end{algorithm}
}

\section{CounterShapley calculation example}
\label{appendix:example}
If we take as an example a counterfactual $\textbf{c}$ and a factual $\textbf{x}$, such that the counterfactual differs from the factual by $5$ feature values:
\begin{equation}
|\delta|=K=5
\end{equation}

Let us iterate through all the possibilities in which these features can be changed \textit{one at a time}, while keeping track of the impact of each change to the prediction score. This is one of three ways in which a Shapley value can be calculated. The example presented here follows the definition in Equation~\ref{eq:shapley}:
\begin{equation*}
  \recallLabel{eq:shapley}
\end{equation*}

Let us start by considering the case where we change two (unspecified) feature values, and then change feature $i$:

\begin{equation}
    [\times\ \times\ i\ \times\ \times]
    \label{eq:xmarksthespot}
\end{equation}

Since $i$ is on the third spot in this example (we change two features, and then we change feature $i$), we have $4$ remaining features we could fill in the first two spots marked as $\times$. This means that the sum over all coalitions will have ${}_{4}C_{2}=6$ elements for the case where $i$ is in position $3$. This is an important result, as it equals the number of model predictions that will have to be made. This means that there are $6$ coalitions $V$ of size $2$ that do not have $i$. This condition appears in the subscript of the summation sign of the Shapley definition in Equation~\ref{eq:shapley}.

Note that, given some coalition of two features that are to be changed before $i$, it does not matter in which order these first two features are changed: both orderings will yield the same marginal contribution for $i$ in position $3$. As we are iterating all order-independent coalitions of size $2$, this iteration step is the only step a particular coalition will be considered for the first two places. To calculate its contribution compared to all possible configurations where feature $i$ is changed as third (including those where another order of the same coalition is considered), we must multiply whatever contribution $i$ in place $3$ yields in this specific configuration by the amount of possible orderings of the precursory coalition: $N_{orders}=|V|! = 2! = 2$. This is done to properly weigh this contribution in the total average over all possible coalitions, and corresponds to the weight of $|V|!$ in Equation~\ref{eq:shapley}.

When we now iterate through all possible subsets of size $2$ that do not have $i$, we can essentially calculate the marginal contribution of feature $i$ when it is changed as a third feature. This is equivalent to the summation sign in Equation~\ref{eq:shapley}, but where the summation would be limited to include only coalitions of size $2$.
Note that it now also does not matter which features will be changed later on. The features that are still to be changed have no influence on the model output difference yielded by changing $i$ as the third feature, which is the only contribution we are counting for now.
The number of iteration steps where $i$ is changed as a third feature change, given that the first two $\times$ are already filled in, can be counted by considering how many features are left to be filled in on the places of the last two $\times$, which equals $(K - |V| - 1)! = (5 - 2 - 1)! = 2! = 2$. Multiplying this with $|V!|$ gives us the total amount of coalitions where $i$ is changed as a third feature (see again Equation~\ref{eq:shapley}), or equivalently, the amount of coalitions of size $2$ that do not contain $i$.

We have now only considered the case where feature $i$ was changed as third.
To gain a complete understanding of the influence of feature $i$, we must also consider the cases where it was changed first, second, fourth, and last. This is equivalent to placing $i$ on each $\times$ spot in Equation~\ref{eq:xmarksthespot}, or to considering coalitions of every possible size that do not have $i$. It is also equivalent to the summation sign in Equation~\ref{eq:shapley}, but now not limited anymore to a fixed coalition size. We can extend the previous logic to all coalition sizes, ranging from $1$ to $5$.

The only thing that remains is to weigh this sum of marginal contributions with the total amount of possible feature permutations by dividing by $K!$. In doing so, we have the average marginal feature contribution of feature $i$, and we have explained the last factor of Equation~\ref{eq:shapley}.

Given a set $\delta$ of size $K$, the total amount of terms Equation \ref{eq:shapley} will have to add together to calculate the Shapley value for a single feature $i$ equals:
\begin{equation}
    N_{coalitions, i} = \sum^{K-1}_{|V|=0} \binom{K-1}{V}=  2^{K-1}-1
\end{equation}
\end{appendices}
\end{document}